\documentclass[accepted]{uai2023} % after acceptance, for a revised
\usepackage[american]{babel}
\usepackage{natbib} % has a nice set of citation styles and commands
\usepackage{mathtools} % amsmath with fixes and additions
\usepackage{booktabs} % commands to create good-looking tables
\usepackage{tikz} % nice language for creating drawings and diagrams
\usepackage{amsfonts} %mathbb
\usepackage{amsmath}
\usepackage{cleveref}
\DeclareMathOperator{\R}{\mathbb R}
\usepackage{xstring}
\usepackage{caption}
\usepackage{subcaption}
\usepackage{amssymb}
\usepackage{amsthm}
\usepackage{bm}
\theoremstyle{plain}
\newtheorem{metric}{Metric}[section]
\newtheorem{theorem}{Theorem}[section]

\newtheorem{lemma}[theorem]{Lemma}
\newtheorem{definition}[theorem]{Definition}

\newcommand{\RomanNumeralCaps}[1]
    {\MakeUppercase{\romannumeral #1}}
    
\newcommand{\appref}[1]{\ref{#1}}

\newcommand\nnfootnote[1]{%
  \begin{NoHyper}
  \renewcommand\thefootnote{}\footnote{#1}%
  \addtocounter{footnote}{-1}%
  \end{NoHyper}
}

\title{A Scalable Walsh-Hadamard Regularizer to Overcome the \\Low-degree Spectral Bias of Neural Networks}

\author[1]{Ali Gorji$^*$}
\author[1]{Andisheh Amrollahi$^*$}
\author[1]{Andreas Krause}
\affil[1]{%
    Computer Science Department\\
    ETH Zurich\\
    Zurich, Switzerland
}

\begin{document}
\maketitle

\begin{abstract}
% \vspace{-4mm}
  % This is the abstract for this article.
  % It should give a self-contained single-paragraph summary of the article's contents, including context, results, and conclusions.
  % Avoid citations; but if you do, you must give essentially the whole reference.
  % For example: This whole paper is devoted to praising É. Š. Åland von Vèreweg's most recent book (“Utopia's government formation problems during the last millenium”, Springevier Publishers, 2016).
  % Also, do not put mathematical notation and abbreviations in your abstract; be descriptive.
  % So not “we solve \(x^2+A xy+y^2\), where \(A\) is an RV”, but “we solve quadratic equations in two unknowns in which a single coefficient is a random variable”.
  % The reason is that mathematical notation will not display correctly when the abstract is reused on the proceedings website, for example, and that one should not assume the abstract's reader knows the abbreviation.
  % Of course the same remarks hold for your paper's title.
\looseness -1 Despite the capacity of neural nets to learn arbitrary functions, models trained through gradient descent often exhibit a bias towards ``simpler'' functions. Various notions of simplicity have been introduced to characterize this behavior. Here, we focus on the case of neural networks with discrete (zero-one), high-dimensional, inputs through the lens of their Fourier (Walsh-Hadamard) transforms, where the notion of simplicity can be captured through the \emph{degree} of the Fourier coefficients. We empirically show that neural networks have a tendency to learn lower-degree frequencies. 
We show how this spectral bias towards low-degree frequencies can in fact \emph{hurt} the neural network's generalization on real-world datasets. To remedy this we propose a new scalable functional regularization scheme that aids the neural network to learn higher degree frequencies. Our regularizer also helps avoid erroneous identification of low-degree frequencies, which further improves generalization. We extensively evaluate our regularizer on synthetic datasets to gain insights into its behavior. Finally, we show significantly improved generalization on four different datasets compared to standard neural networks and other relevant baselines.   
\end{abstract}
\nnfootnote{*These authors contributed equally to this work}

% \vspace{-5mm}
\section{Introduction}\label{sec:intro}
% \vspace{-3mm}
% First paragraph
% \begin{enumerate}
%     \item Importance of regularization
% \end{enumerate}

\looseness -1 Classical work on neural networks shows that deep fully connected neural networks have the capacity to approximate arbitrary functions \citep{hornik1989multilayer, cybenko1989approximation}. However, in practice, neural networks trained through (stochastic) gradient descent have a ``simplicity'' bias. This notion of simplicity is not agreed upon and works such as \citep{arpit_closer_2017, nakkiran2019sgd, valle2018deep, kalimeris_sgd_2019} each introduce a different notion of ``simplicity''. The simplicity bias can also be studied by considering the function the neural net represents (function space view) and modeling it as Gaussian processes (GP)\citep{rasmussen2004gaussian}. \citet{daniely2016toward, lee2018deep} show that a wide, randomly initialized, neural network in function space is a sample from a GP  with a kernel called the ``Conjugate Kernel'' \citep{daniely2017sgd}. Moreover, the evolution of gradient descent on a randomly initialized neural network  can be described through the ``Neural Tangent Kernel'' \citet{jacot_neural_2018, lee2019wide}. These works open up the road for analyzing the simplicity bias of neural nets in terms of a \emph{spectral} bias in Fourier space. \citet{rahaman_spectral_2019} show empirically that neural networks tend to learn sinusoids of lower frequencies earlier on in the training phase compared to those of higher frequencies. Through the GP perspective introduced by  \citet{jacot_neural_2018, lee2019wide}, among others, \citet{ronen_convergence_2019, basri2020frequency} were able to prove these empirical findings. These results focus on {\em continuous} domains and mainly emphasize the case where the input and output are both \emph{one-dimensional}. 

Here,  we focus on {\em discrete} domains where the input is a \emph{high-dimensional} zero-one vector and we analyze the function learned by the neural network in terms of the amount of interactions among its input features in a quantitative manner. Our work is complementary to the majority of the aforementioned work that has been done on the spectral bias of neural networks in the setting of \emph{continuous}, \emph{one-dimensional} inputs  \citep{ronen_convergence_2019, basri2020frequency, rahaman_spectral_2019}. \citet{yang_fine-grained_2020, valle2018deep}  are the first to provide spectral bias results for the discrete, higher dimensional, setting (our setting). By viewing a fully connected neural network as a function that maps zero-one vectors to real values, one can expand this function in terms of the Fourier --a.k.a Walsh-Hadamard -- basis functions. The Walsh-Hadamard basis functions have a natural ordering in terms of their complexity called their {\em degree}. The degree specifies how many features each basis function is dependent upon. For example, the zero-degree basis function is the constant function and the degree-one basis functions are functions that depend on exactly one feature. Through analysis of the NTK gram matrix on the Boolean cube, \citet{yang_fine-grained_2020} theoretically show that, roughly speaking, neural networks learn the lower degree basis functions earlier in training. 

% Here we extend their work by analyzing a neural network trained on summations of such functions i.e $k>1$-sparse functions and analyzing their behavior during training (over different epochs). We also checked for a variety of different dataset sizes. Our ultimate goal is to improve their generalization performance in learning these functions.  Therefore, we conduct a variety of experiments that shed insight into neural networks' degree-dependent, spectral behavior during training and its effect on overfitting and generalization. This guides us in proposing our novel regularizer \textsc{HashWH}. 

\looseness -1 This tendency to prioritize simpler functions in neural networks has been suggested as a cardinal reason for their remarkable generalization ability despite their over-parameterized nature \citep{neyshabur_exploring_2017, arpit_closer_2017, kalimeris_sgd_2019, poggio_theory_2018}. However, much less attention has been given to the case where the simplicity bias can {\em hurt} generalization \citep{tancik2020fourier, shah_pitfalls_2020}. 
%\citet{shah_pitfalls_2020} show that neural networks after training are functionally dependent on the ``simplest'' feature, even if it is less predictive than more complex features, in terms of for example classification accuracy. The features in their work is continuous and a ``simple'' feature is one that solves the problem of classification with single thresholding as opposed to several thresholds. 
\citet{tancik2020fourier} show how transforming the features with random Fourier features embedding helps the neural network overcome its spectral bias and achieve better performance in a variety of tasks. They were able to explain, in a unified way, many empirical findings in computer vision research such as sinusoidal positional embeddings through the lens of overcoming the spectral bias. In the same spirit as these works, we show that the spectral bias towards low-degree functions  can hurt generalization and how to remedy this through our proposed regularizer.

%\citep{yang_fine-grained_2020, ronen_convergence_2019, valle2018deep, neyshabur2015search, neyshabur_exploring_2017-1, soudry_implicit_2018, poggio_theory_2018, arpit_closer_2017, kalimeris_sgd_2019, rajeswaran_towards_2017, huh_low-rank_2022}

%were introduced to prevent overfitting to the training data or equivalently enhance generalizability. Weight decay or L2-regularization,
%\citep{krogh1991simple},
%L1-regularization, batch normalization,
% \citep{ioffe2015batch}, 
%and dropout
% \citep{srivastava2014dropout}
%are prominent examples that are widely adopted. 
In more recent lines of work, regularization schemes have been proposed to directly impose priors on the function the neural network represents \citep{benjamin_measuring_2019, sun_functional_2019, wang_function_2019}. This is in contrast to other methods such as dropout, batch normalization, or other methods that regularize the weight space. 
In this work, we also regularize neural networks in function space by imposing sparsity constraints on their Walsh-Hadamard transform. Closest to ours is the work of \citet{aghazadeh_epistatic_2021}. Inspired by studies showing that biological landscapes are sparse and contain high-degree frequencies \citep{sailer_detecting_2017, yang_higher-order_2019, brookes_sparsity_2022, ballal_sparse_2020, poelwijk_learning_2019}, they propose a functional regularizer to enforce sparsity in the Fourier domain and report improvements in generalization scores.
% , ,,, , , , }).
% * Worth rethinking candidates:
% -- pan_continual_2020: not the vanilla NN setup but the Continual Learning, an online learning setup in which tasks are presented sequentially. But literally regularizing the network in function space using GPs (https://proceedings.neurips.cc/paper/2020/file/2f3bbb9730639e9ea48f309d9a79ff01-Paper.pdf).
% -- cheng_control_2019: does the functional regularisation over three famous Deep RL frameworks (https://arxiv.org/pdf/1905.05380.pdf)
% -- piche_bridging_2022: seems to be really functional regularizing the value network in Deep RL, preprint though (https://arxiv.org/pdf/2210.12282.pdf).
% -- lyle_understanding_2022: functional regularization again in deep RL settings, "we proposed a regularizer to preserve capacity, yielding improved performance across a number of settings in which deep RL agents have historically struggled to match human performance.", (https://openreview.net/forum?id=ZkC8wKoLbQ7).

% * Probably irrelevant but still function space:
% -- titsias_functional_2020: "We introduced a functional regularisation approach for supervised continual learning that combines inducing point GP inference with deep neural networks.", good paper but not regularising Neural networks
% -- von_oswald_continual_2022: Continual learning setup + weight generator for NN
% -- li_fast_2019: despite directly working with Funtion space, defined over GNNs as well as being an operator than a regularizer
% rudner_tractable_2022: BNN, not regular NNs

\textbf{Our contributions:}
% \vspace{-2mm}
\begin{itemize}[leftmargin=*]
    \item We analyze the spectral behavior of a simple MLP during training through extensive experiments. We show that the standard (unregularized) network not only is unable to learn (more complex) high-degree frequencies but it also starts learning erroneous low-degree frequencies and hence overfitting on this part of the spectrum.  
    \item \looseness -1 We propose a novel regularizer -- \textsc{HashWH} (Hashed Walsh Hadamard) -- to remedy the aforementioned phenomenon. The regularizer acts as a ``sparsifier'' on the Fourier (Walsh-Hadamard) basis. In the most extreme cases, it reduces to simply imposing an $L_1$-norm on the Fourier transform of the neural network. Since computing the exact Fourier transform of the neural net is intractable, our regularizer hashes the Fourier coefficients to buckets and imposes an L1 norm on the buckets. By controlling the number of hash buckets, it offers a smooth trade-off between computational complexity and the quality of regularization. 
    \item We empirically show that \textsc{HashWH} aids the neural network in avoiding erroneous low-degree frequencies and also learning relevant high-degree frequencies. %The regularizer guides gradient descent to train a function that has less energy in the lower-degree components and more energy in the high-degree components of its spectrum. 
    The regularizer guides the training procedure to allocate more energy to the high-frequency part of the spectrum when needed and allocate less energy to the lower frequencies when they are not present in the dataset.
    
    \item We show on real-world datasets that, contrary to popular belief of simplicity biases for neural networks, fitting a low degree function does not imply better generalization. Rather, what is more important, is keeping the \emph{higher amplitude} coefficients regardless of their degree. We use our regularizer on four real-world datasets and provide state of the art results in terms of $R^2$ score compared to standard neural networks and other baseline ML models, especially for the low-data regime. %We plot its performance for a variety of different training set sizes. 
\end{itemize}

\section{Background}\label{sec:background}
% \vspace{-3mm}
In this section, we first review Walsh Hadamard transforms, and notions of degree and sparsity in the Fourier (Walsh-Hadamard) domain \citep{o2014analysis}. Next, we review the notion of simplicity biases in neural networks and discuss why they are spectrally biased toward low-degree functions.
% \vspace{-7mm}
\subsection{Walsh Hadamard transforms}
\label{subsec:wht_background}
% \vspace{-3mm}
Let $g:\{0,1\}^n \rightarrow \R$ be a function mapping Boolean zero-one vectors to the real numbers, also known as a ``pseudo-boolean'' function. The family of $2^n$ functions $\{\Psi_f: \{0,1\}^n \rightarrow \R | f \in \{0,1\}^n\}$ defined below consists of the Fourier basis functions. This family forms a basis over the vector space of all pseudo-boolean functions:
\[
\Psi_f(x) = \frac{1}{\sqrt{2^n}} (-1)^{\langle f, x \rangle}, f,x \in \{0,1\}^n
\]
where ${\langle f, x \rangle} = \sum_i f_i x_i$. Here, $f \in \{0,1\}^n$ is called the \emph{frequency} of the basis function. For any frequency $f \in \{0,1\}^n$ we denote its \emph{degree} by $\text{deg}(f)$  which is defined as the number of non-zero elements. For example, $f_1=[0,0,0,0,0]$ and $f_2=[0,0,1,0,1]$ have degrees $\text{deg}(f_1)=0$ and $\text{deg}(f_2)=2$, respectively. One can think of the degree as a measure of the complexity of basis functions. For example, $\Psi_0(x)$ is constant, and $\Psi_{e_i}(x)$ where $e_i$ is a standard basis vector ($\text{deg}(e_i)=1$) only depends on feature $i$ of the input. It is equal to $+1$ when feature $i$ is zero and equal to $-1$ when feature $i$ is one. More generally, a degree $d$ basis function depends on exactly $d$ input features.  

Since the Fourier basis functions form a basis for the vector space of all pseudo-boolean functions, any function $g:\{0,1\}^n \rightarrow \R $ can be written as a unique linear combination of these basis functions:
\[
g(x) =  \frac{1}{\sqrt{2^n}} \sum\limits_{f \in \{0,1\}^n}  \widehat{g}(f)  (-1)^{\langle f, x \rangle}
\]
The (unique) coefficients $\widehat{g}(f)$ are called the ``Fourier coefficients'' or ``Fourier amplitudes'' and are computed as $
\widehat{g}(f) =  \frac{1}{\sqrt{2^n}} \sum\limits_{x \in \{0,1\}^n}  g(x)  (-1)^{\langle f, x \rangle}
$. 
The \emph{Fourier spectrum} of $g$ is the vector consisting of all of its $2^n$ Fourier coefficients, which we denote by the bold symbol $\widehat{\mathbf{g}}\in\mathbb{R}^{2^n}$. Assume $\mathbf{X}\in \{0,1\}^{2^n\times n}$ to be the matrix of an enumeration over all possible $n$-dimensional binary sequences ($\{0,1\}^n$), and $\mathbf{g}(\mathbf{X})\in \mathbb{R}^{2^n}$ to be the vector of $g$ evaluated on the rows of $\mathbf{X}$. We can compute the Fourier spectrum using Walsh-Hadamard transform as $
\widehat{\mathbf{g}} = \frac{1}{\sqrt{2^n}}\mathbf{H}_n \mathbf{g}(\mathbf{X})$, 
where $\mathbf{H}_n \in \{\pm1\}^{2^n\times 2^n}$ is the orthogonal Hadamard matrix (see Appendix~\appref{app:sec:walsh_hadamard}).

Lastly, we define the \emph{support} of $g$ as the set of frequencies with non-zero Fourier amplitudes $\text{supp}(g) := \{f \in \{0,1\}^n | \widehat{g}(f) \neq 0\}$. The function $g$ is called \emph{$k$-sparse} if $|\text{supp}(g)| \leq k $. The function $g$ is called \emph{of degree} $d$ if all frequencies in its support have degree at most $d$. 
% \vspace{-5mm}
\subsection{Spectral bias theory} 
% \vspace{-3mm}
The function that a ReLU neural network represents at initialization can be seen as a sample from a GP  $N(0,K)$ in the infinite width limit \citep{daniely2016toward, lee2018deep} (randomness is over the initialization of the weights and biases). The kernel $K$ of the GP is called the ``Conjugate Kernel'' \citep{daniely2016toward} or the ``nn-GP kernel'' \citep{lee2018deep}. Let the kernel Gram matrix $\mathcal{K}$ be formed by evaluating the kernel on the Boolean cube i.e. $\{0,1\}^n$ and let $\mathcal{K}$ have the following spectral decomposition: $\mathcal{K} = \sum\limits_{i=1}^{2^n} \lambda_i u_i u_i^\top$, where we assume that the eigenvalues $\lambda_1 \geq \dots \geq \lambda_{2^n}$ are in decreasing order. 
Each sample of the GP can be obtained as $
\sum\limits_{i=1}^{2^n} \lambda_i \bm{w_i} u_i, \bm{w_i} \sim \mathcal{N}(0,1)$. 
Say that $\lambda_1 \gg \sum_{i \geq 2} \lambda_i$. Then a sample from the GP will, roughly speaking, look very much like $u_1$. 

Let $u_f, f\in\{0,1\}^n$ be obtained by evaluating the Fourier basis function $\Psi_f$ at the $2^n$ possible inputs on $\{0,1\}^n$. \citet{yang_fine-grained_2020} show that $u_f$ is a eigenvector for $\mathcal{K}$. Moreover, they show (weak) spectral bias results in terms of the degree of $f$. Namely, the eigenvalues corresponding to higher degrees have smaller values \footnote{To be more precise, they show that the eigenvalues corresponding to even and odd degree frequencies form decreasing sequences. That is, even and odd degrees are considered separately.}. The result is \emph{weak} as they do not provide a \emph{rate} as to which the eigenvalues decrease with increasing degrees. Their results show that neural networks are similar to low-degree functions at initialization. 

Other works show that in infinite-width neural networks weights after training via (stochastic) gradient descent do not end up too far from the initialization \citep{chizat2019lazy, jacot_neural_2018, du2018gradient, allen2019convergence, allen2019convergencernn}, referred to as ``lazy training'' by \citet{chizat2019lazy}. \citet{lee2018deep, lee2019wide} show that training the last layer of a randomly initialized neural network via full batch gradient descent for an infinite amount of time corresponds to GP posterior inference with the kernel $K$. \citet{jacot_neural_2018, lee2019wide} proved that when training \emph{all} the layers of a neural network (not just the final layer), the evolution can be described by a kernel called the ``Neural Tangent Kernel'' and the trained network yields the mean prediction of GP $N(0, K_{NTK})$ \citep{yang_fine-grained_2020} after an infinite amount of time. \citet{yang_fine-grained_2020} again show that $u_f$ are eigenvectors and weak spectral bias holds. Furthermore, \citet{yang_fine-grained_2020} provides empirical results for the generalization of neural nets of different depths on datasets arising from $k=1$-sparse functions of varying degrees.

\section{Low-degree spectral bias}\label{sec:method}
% \vspace{-3mm}
%For better readability, in this section, we refer to different regularization techniques by their name, and the Neural Network without any further regularizations as \emph{standard}.

In this section, we conduct experiments on synthetically generated datasets to show neural networks' spectral bias and their preference toward learning lower-degree functions over higher-degree ones. Firstly, we show that the neural network is not able to pick up the high-degree frequency components. Secondly, it can learn erroneous lower-degree frequency components. To address these issues, in Section \ref{sec:en}, we introduce our regularization scheme called \emph{\textsc{HashWH}} (Hashed Walsh Hadamard) and demonstrate how it can remedy both problems.

% The experiments are conducted in two different settings. One in a relatively low-dimension space to be able to finely monitor the behavior of different regularization schemes, and one in higher dimensions to highlight the power of our regularization method in real-world problems both in terms of computational complexity and also sparsifying the spectrum. 
\subsection{Fourier spectrum evolution} 
\begin{figure*}[!htb]
    \centering
    \begin{subfigure}[b]{0.75\linewidth}
         \centering
         \includegraphics[width=\linewidth]{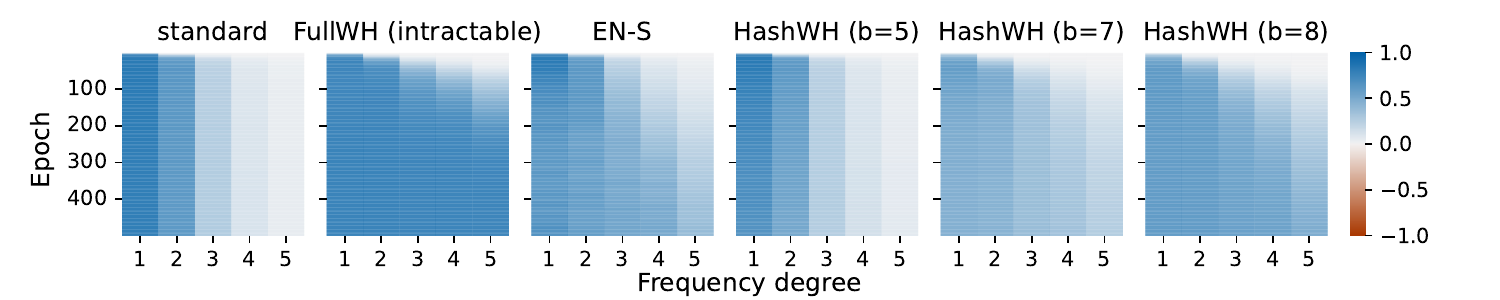}
         \caption{Target support}
         \label{fig:data_freq_heatmap}
     \end{subfigure}

     \begin{subfigure}[b]{0.9\linewidth}
         \centering
         \includegraphics[width=\linewidth]{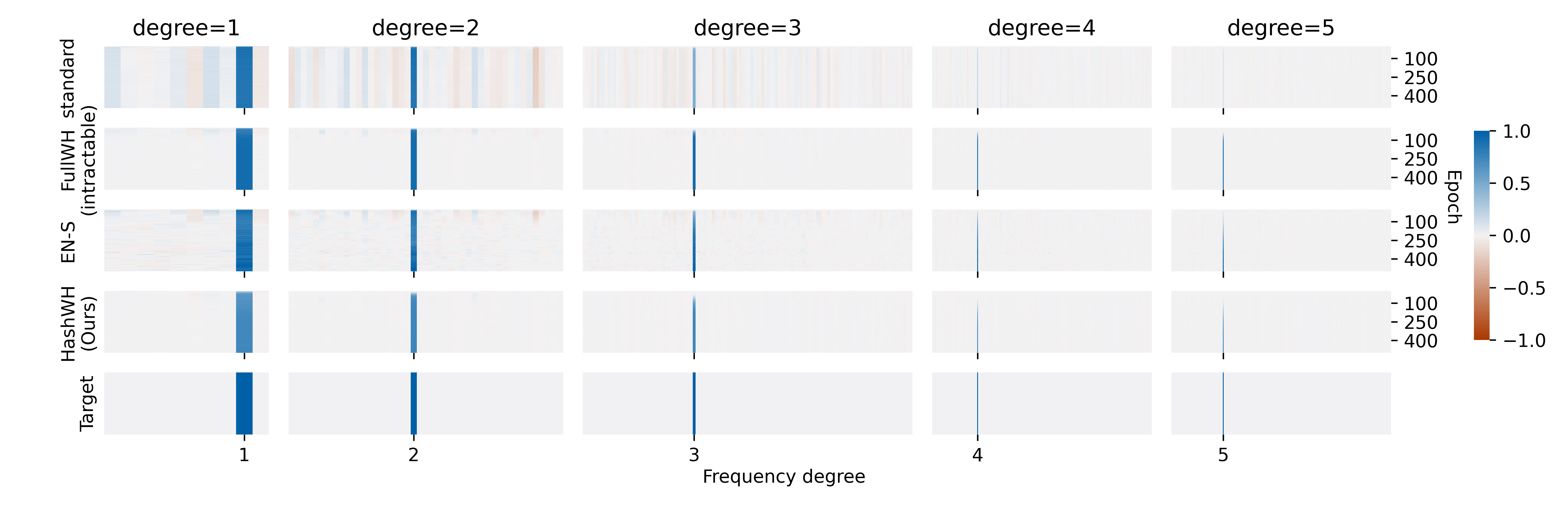}
         \caption{Whole Fourier spectrum}
         \label{fig:sc_heatmap}
     \end{subfigure}
    \caption{Evolution of the Fourier spectrum during training. \textsc{Standard} is the unregularized neural network. \textsc{FullWH} imposes $L_1$-norm regularization on the exact Fourier spectrum and is intractable. \textsc{EN-S} alternates between computing a sparse Fourier approximation (computationally very expensive) and regularization. \textsc{HashWH} (ours) imposes $L_1$ regularization on the hashed spectrum. Figure (a) is limited to the target support. The standard neural network is unable to learn higher degree frequencies. Our regularizer fixes this. Figure (b) is on the whole spectrum. The standard neural network picks up erroneous low-degree frequencies while not being able to learn the higher-degree frequencies. Our regularizer fixes both problems.}
\end{figure*}

We analyze the evolution of the function learned by neural networks during training. We train a neural network on a dataset arising from  a synthetically generated sparse function with a low-dimensional input domain. Since the input is low-dimensional it allows us to calculate the Fourier spectrum of the network (exactly) at the end of each epoch.
% We compare the evolution of the Fourier spectrum of Neural Networks trained with and without sparsity regularization. Our regularizations of interest are EN, EN-S with $m=4$, and \textsc{HashWH} with $b\in\{5, 7, 8\}$. For each model, we compute the output of the network on all $2^{10}$ possible inputs after each training epoch using which we compute the Fourier spectrum of the network.

\textbf{Setup.} 
Let $g^*:\{0, 1\}^{10} \rightarrow \mathbb{R}$ be a synthetic function with five frequencies in its support with degrees 1 to 5 ($\text{supp}(g^*)=\{f_1, f_2, f_3, f_4, f_5\}, \text{deg}(f_i)=i$), all having equal Fourier amplitudes of $\widehat{g}^*(f_i)=1$. Each $f_i$ is sampled uniformly at random from all possible frequencies of degree $i$. The training set is formed by drawing uniform samples from the Boolean cube $x\sim \mathcal{U}_{\{0,1\}^{10}}$ and evaluating $g^*(x)$.

% , i.e., random initialization of the network, random samples used in EN-S for sparse Fourier spectrum approximation, and sampling of hashing matrices in our method

We draw five such target functions $g^*$ (with random support frequencies). For each draw of the target function, we create five different datasets all with 200 training points and sampled uniformly from the input domain but with different random seeds. We then train a standard five-layer fully connected neural network using five different random seeds for the randomness in the training procedure (such as initialization weights and SGD). We aggregate the results over the $125$ experiments by averaging.
%to reduce the variance resulting from the randomness in the choice of $\text{supp}(g^*)$, training data, and training randomness. 
We experiment the same setting with three other training set sizes.
Results with training set size other than 200 and further setup details are reported in Appendices~\appref{app:subsec:evolution_detailed} and \appref{app:sec:technical_details}, respectively. % In all that follows, we refer to the $g^*$ used to generate each dataset as \emph{target} function, and the frequencies in its support as the target support.

\textbf{Results.} 
We first inspect the evolution of the learned Fourier spectrum over different epochs and limited to the target support ($\text{supp}(g^*)$). Figure~\ref{fig:data_freq_heatmap} shows the learned amplitudes for frequencies in the target support at each training epoch.
%averaged over the 125 aforementioned runs.
% (made possible due to always having a single frequency for each degree).
Aligned with the literature on simplicity bias \citep{valle2018deep,yang_fine-grained_2020}, we observe that neural networks learn the low-degree frequencies earlier in the epochs. Moreover, we can see in the left-most figure in Figure~\ref{fig:data_freq_heatmap} that despite eventually learning low-degree frequencies, the standard network is unable to learn high-degree frequencies.
% However, all sparsity-inducing regularization methods display better performance in learning high-degree frequencies. To the extent that EN is capable of perfectly learning all target frequencies. It can also be seen that increasing the size of the hashing matrix in \textsc{HashWH} boosts the learning of high-degree frequencies.

Next, we expand the investigation to the whole Fourier spectrum instead of just focusing on the support frequencies. The first row of Figure~\ref{fig:sc_heatmap} shows the evolution of the Fourier spectrum during training and compares it to the spectrum of the target function on the bottom row. We average the spectrum linked to one of the five target synthetic functions (over the randomness of the dataset sampling and training procedure) and report the other four in Appendix~\appref{app:subsec:evolution_detailed}. We observe that in addition to the network not being able to learn the high-degree frequencies, the standard network is prone to learning incorrect low-degree frequencies as well. 
%This is another artifact of the simplicity bias. 

% It can be seen that besides the better performance of the sparsity-inducing methods in learning the target frequencies, they are also better at filtering non-relevant frequencies. The standard model, however, has wrongly learned multiple low-degree frequencies over the course of training.

% \vspace{-3mm}
% \section{Sparsifying the Fourier spectrum} 
\section{Overcoming the spectral bias via regularization}
\label{sec:en}
% \vspace{-2mm}
Now, we introduce our regularization scheme \emph{\textsc{HashWH}} (Hashed Walsh-Hadamard). 
Our regularizer is essentially a ``sparsifier'' in the Fourier domain. That is, it guides the neural network to have a sparse Fourier spectrum. We empirically show later how sparsifying the Fourier spectrum can both stop the network from learning erroneous low-degree frequencies and aid it in learning the higher-degree ones, hence remedying the two aforementioned problems.

Assume $\mathcal{L}_{net}$ is the loss function that a standard neural network minimizes, e.g., the MSE loss in the above case.  We modify it by adding a regularization term $\lambda \mathcal{L}_{sparsity}$. Hence the total loss is given by: $\mathcal{L} = \mathcal{L}_{net} + \lambda \mathcal{L}_{sparsity}$.
% The sparsity loss $\mathcal{L}_{sparsity}$ should be lower when the Fourier spectrum of the network, $\widehat{g_\theta}$, is sparser. 

The most intuitive choice is $\mathcal{L}_{sparsity}=\|\widehat{\mathbf{g_N}}\|_0$, where $\widehat{\mathbf{g_N}}$ is the Fourier spectrum of the neural network function $g_N: \{0,1\}^n \rightarrow \R$. Since the $L_0$-penalty's derivative is zero almost everywhere, one can use its tightest convex relaxation, the $L_1$-norm, which is also sparsity-inducing, as a surrogate loss. \citet{aghazadeh_epistatic_2021} use this idea and name it as Epistatic-Net or ``EN'' regularization: $\mathcal{L}_{EN} := \mathcal{L}_{net} + \lambda \|\widehat{\mathbf{g_N}}\|_1\label{eq:EN_loss}$. In this work, we call this regularization \textsc{FullWH} (Full Walsh Hadamard transform).
% EN requires the computation of the network output on all $2^n$ possible inputs at each iteration of the back-propagation. Therefore, the computational complexity grows \emph{exponentially} with the number of dimensions $n$, making it computationally intractable for $n>20$. To avoid the burden of computing network output on all possible inputs, we employ a hashing technique to approximate the $\|\widehat{\mathbf{g_\theta}}\|_1$ term.

\textsc{FullWH} requires the evaluation of the network output on all $2^n$ possible inputs at each iteration of back-prop. Therefore, the computational complexity grows \emph{exponentially} with the number of dimensions $n$, making it computationally intractable for $n>20$ in all settings of practical importance.

\looseness -1 \citet{aghazadeh_epistatic_2021} also suggest a more scalable version of \textsc{FullWH}, called ``\textsc{EN-S}'',
%by considering  $\widehat{\mathbf{g_\theta}}$ as an explicit optimization constraint and decoupling the optimization of $\mathcal{L}_{EN}$ into two separate minimization problems of network optimization and Fourier spectrum sparsification followed by a dual update, using ADMM (\citet{boyd_distributed_2011}), which enabled them to iteratively approximate a sparse Fourier spectrum for the network at each epoch. 
which roughly speaking, alternates between computing the sparse \emph{approximate} Fourier transform of the network at the end of each epoch and doing normal back-prop, as opposed to the exact computation of the exact Fourier spectrum when back-propagating the gradients. 
In our experiments, we show \textsc{EN-S} can be computationally expensive because the sparse Fourier approximation primitive can be time-consuming. For a comprehensive comparison see Appendix \appref{app:subsec:EN-S}. Later, we show that empirically, it is also less effective in overcoming the spectral bias as measured by achievable final generalization error.

%where $m$ is  the hash size of the Fourier approximation algorithm used.

% \vspace{-5mm}
\subsection{\textsc{HashWH}}
\label{subsec:Hashwh}
% \vspace{-3mm}
We avoid the exponentially complex burden of computing the exact Fourier spectrum of the network by employing a hashing technique to approximate the regularization term $\lambda \|\widehat{\mathbf{g_N}}\|_1$.
Let $g:\{0, 1\}^n \rightarrow \mathbb{R}$ be a pseudo-boolean function. We define the lower dimensional function $u_{\mathbf{\sigma}}: \{0, 1\}^b \rightarrow \mathbb{R}$, where $b \ll n$,  by sub-sampling $g$ on its domain: $u_\mathbf{\sigma}(\tilde{x}) \triangleq \sqrt{\frac{2^n}{2^b}} \ g(\mathbf{\sigma} \tilde{x}), \ \tilde{x} \in \{0,1\}^b$
where $\mathbf{\sigma} \in \{0,1\}^{n \times b}$ is some matrix which we call the \emph{hashing matrix}. The matrix-vector multiplication $\sigma \tilde{x}$is taken modulo 2. $u_\sigma$ is defined by sub-sampling $g$ on all the points lying on the (at most) $b$-dimensional subspace spanned by the columns of the hashing matrix $\sigma$. The special property of sub-sampling the input space from this subspace is in the arising Fourier transform of $u_\sigma$ which we will explain next. 

The Fourier transform of $u_{\mathbf{\sigma}}$ can be derived as (see Appendix~\appref{app:subsec:hash_sum_proof}):
\begin{align}
    \widehat{u}_\mathbf{\sigma}(\Tilde{f}) = \sum_{f \in \{0,1\}^n:\ \mathbf{\sigma}^\top f= \Tilde{f}} \widehat{g}(f), \ \Tilde{f} \in \{0,1\}^b
    \label{eq:hash_sum}
\end{align}
% \todo{Prove it in Appendix}
One can view $\widehat{u}_\mathbf{\sigma}(\Tilde{f})$ as a ``bucket'' containing the sum of all Fourier coefficients $\widehat{g}(\tilde{f})$ that are ``hashed'' (mapped) into  it by the linear hashing function $h(f) = \sigma^\top f$. There are $2^b$ such buckets and each bucket contains frequencies lying in the kernel (null space) of the hashing map plus some shift. 
% , the Fourier transform of which on an arbitrary bucket is the sum of original Fourier coefficients of frequencies hashed into it. 

In practice, we let $\mathbf{\sigma} \sim \mathcal{U}_{\{0,1\}^{n \times b}}$ be a uniformly sampled hash matrix that is re-sampled after each iteration of back-prop. Let $\mathbf{X}_b \in \{0, 1\}^{2^b \times b}$ be a matrix containing as rows the enumeration over all points on the Boolean cube $\{0, 1\}^b$. Our regularization term approximates \eqref{eq:EN_loss} and is given by:
\begin{equation}
    \mathcal{L}_{\textsc{HashWH}} 
    \triangleq \mathcal{L}_{net} + \lambda \| \mathbf{H}_b \mathbf{g_N}(\mathbf{X}_b \mathbf{\sigma}^T) \|_1
    = \mathcal{L}_{net} + \lambda \|\widehat{\mathbf{u_\mathbf{\sigma}}}\|_1 \nonumber 
    \label{eq:HashWH_loss}
\end{equation}
\looseness -1 That is, instead of imposing the $L_1$-norm directly on the whole spectrum, this procedure imposes the norm on the ``bucketed'' (or partitioned) spectrum where each bucket (partition) contains sums of coefficients mapped to it. The larger $b$ is the more partitions we have and the finer-grained the sparsity-inducing procedure is. Therefore, the quality of the approximation can be controlled by the choice of $b$. Larger $b$ allows for a finer-grained regularization but, of course, comes at a higher computational cost because a Walsh-Hadamard transform is computed for a higher dimensional sub-sampled function $u$. Note that $b=n$ corresponds to hashing to $2^n$ buckets. As long as the hashing matrix is invertible, this precisely is the case of \textsc{FullWH} regularization. 

The problem with the above procedure arises when, for example, two ``important'' frequencies $f_1$ and $f_2$ are hashed into the same bucket, i.e., $\mathbf{\sigma}^\top f_1 = \mathbf{\sigma}^\top f_2$, an event which we call a ``collision''. This can be problematic when the absolute values $|\widehat{g}(f_1)|$ and $|\widehat{g}(f_2)|$ are large (hence they are important frequencies) but their sum can cancel out due to differing signs. In this case, the hashing procedure can zero out the sum of these coefficients. We can reduce the probability of a collision by increasing the number of buckets, i.e., increasing $b$ \citep{alon1999linear}. 

In Appendix~\appref{app:subsec:collision_probability} we show that the expected number of collisions $C$ is given by:
$\mathbb{E}[C]=\frac{(k-1)^2}{2^{b}}$ which decreases linearly with  the number of buckets $2^b$. Furthermore, we can upper bound the probability $p$ that a given important frequency $f_i$ collides with any other of the $k-1$ important frequencies in one round of hashing. Since we are independently sampling a new hashing matrix $\mathbf{\sigma}$ at each round of back-prop, the number of collisions of a given frequency over the different rounds has a binomial distribution. In Appendix~\appref{app:subsec:collision_probability} we show that picking $b \geq \log_2(\frac{k-1}{\epsilon}), \epsilon>0$ guarantees that collision of a given frequency happens approx.~an $\epsilon$-fraction of the $T$ rounds, and not much more.

\textbf{Fourier spectrum evolution of different regularization methods.} 
We analyze the effect of regularizing the network with various Fourier sparsity regularizers in the setting of the previous section. 
%monitoring the evolution of the Fourier spectrum using the same datasets and 
%random seeding strategy. 
Our regularizers of interest are \textsc{FullWH}, \textsc{EN-S} with $m=5$ ($2^m$ is the number of buckets their sparse Fourier approximation algorithm hashes into), and \textsc{HashWH} with $b \in \{5, 7, 8\}$.  

Returning to Figure~\ref{fig:data_freq_heatmap}, we see that despite the inability of the standard neural network in picking up the \emph{high-degree} frequencies, all sparsity-inducing regularization methods display the capacity for learning them. \textsc{FullWH} is capable of perfectly learning the entire target support. It can also be seen that increasing the size of the hashing matrix in \textsc{HashWH} (ours) boosts the learning of high-degree frequencies. Furthermore, Figure~\ref{fig:sc_heatmap} shows that in addition to the better performance of the sparsity-inducing methods in learning the target support, they are also better at filtering out non-relevant \emph{low-degree} frequencies. 
% This suggests that enforcing a sparse Fourier spectrum during training can help the network learn relevant high-degree frequencies.

We define a notion of approximation error which is basically the normalized energy of the error in the learned Fourier spectrum on an arbitrary subset of frequencies.
% \vspace{-1mm}
\begin{metric}[Spectral Approximation Error (SAE)]
Let $g_N: \{0, 1\}^n\rightarrow \mathbb{R}$ be an approximation of the target function $g^*: \{0, 1\}^n\rightarrow \mathbb{R}$. Consider a subset of frequencies $S \subseteq \{0,1\}^n$, and assume $\widehat{\mathbf{g_N}}_S$ and $\widehat{\mathbf{g^*}}_S$ to be the vector of Fourier coefficients of frequencies in $S$, for $g_N$ and $g^*$ respectively. As a measure of the distance between $g_N$ and $g$ on the subset of frequencies $S$, we define Spectral Approximation Error as: $
    \text{SAE} = \frac{\|\widehat{\mathbf{g_N}}_S - \widehat{\mathbf{g^*}}_S\|^2_2}{\|\widehat{\mathbf{g^* }}_S\|^2_2} % \label{eq:function_error}
$
% which is the normalized energy of the difference over frequencies in $S$. In the case of $S=\{0,1\}^n$, Function Error is the normalized energy of the difference in the whole Fourier spectrum.
% \vspace{-4mm}
\end{metric}
\begin{figure}
  \centering
  \includegraphics[width=\linewidth]{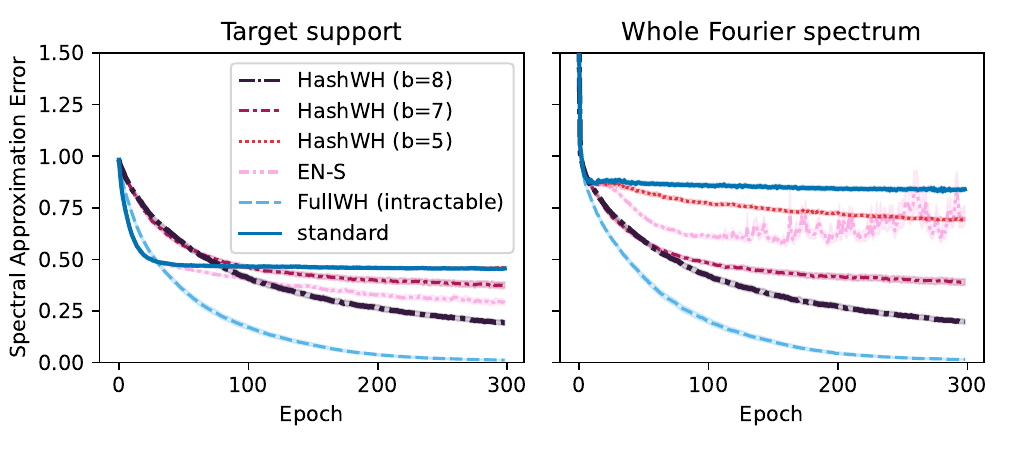}
  % \vspace*{-8mm}
  \caption{\looseness -1 Evolution of the spectral approximation error during training. The left plot limits the error to the target support, while the right one considers the whole Fourier spectrum. For the standard neural network, the SAE is considerably worse on the full spectrum which shows the importance of eliminating the erroneous frequencies that are not in the support of the target function. We also see the graceful scaling of SAE of \textsc{HashWH} (ours) with the hashing matrix size.}
  % \vspace*{0mm}
  \label{fig:function_error}
\end{figure}

Figure \ref{fig:function_error} shows the SAE of the trained network using different regularization methods over epochs, for both when $S$ is target support as well as when $S=\{0,1\}^n$ (whole Fourier spectrum). 
% The standard network achieves an acceptable SAE but is still worse than \textsc{HashWH} (ours) on the target support (in all cases). However, the SAE is considerably worse for the standard neural net when taken on the full spectrum. 
The standard network displays a significantly higher (worse) SAE on the whole Fourier spectrum compared to the target support, while Walsh-Hadamard regularizers exhibit consistent performance across both.
This shows the importance of enforcing the neural network to have zero Fourier coefficients on the non-target frequencies. Moreover, we can see \textsc{HashWH} (ours) leads to a reduction in SAE that can be smoothly controlled by the size of its hashing matrix. 

\begin{figure*}[!htb]
  \centering
  \includegraphics[width=0.9\linewidth]{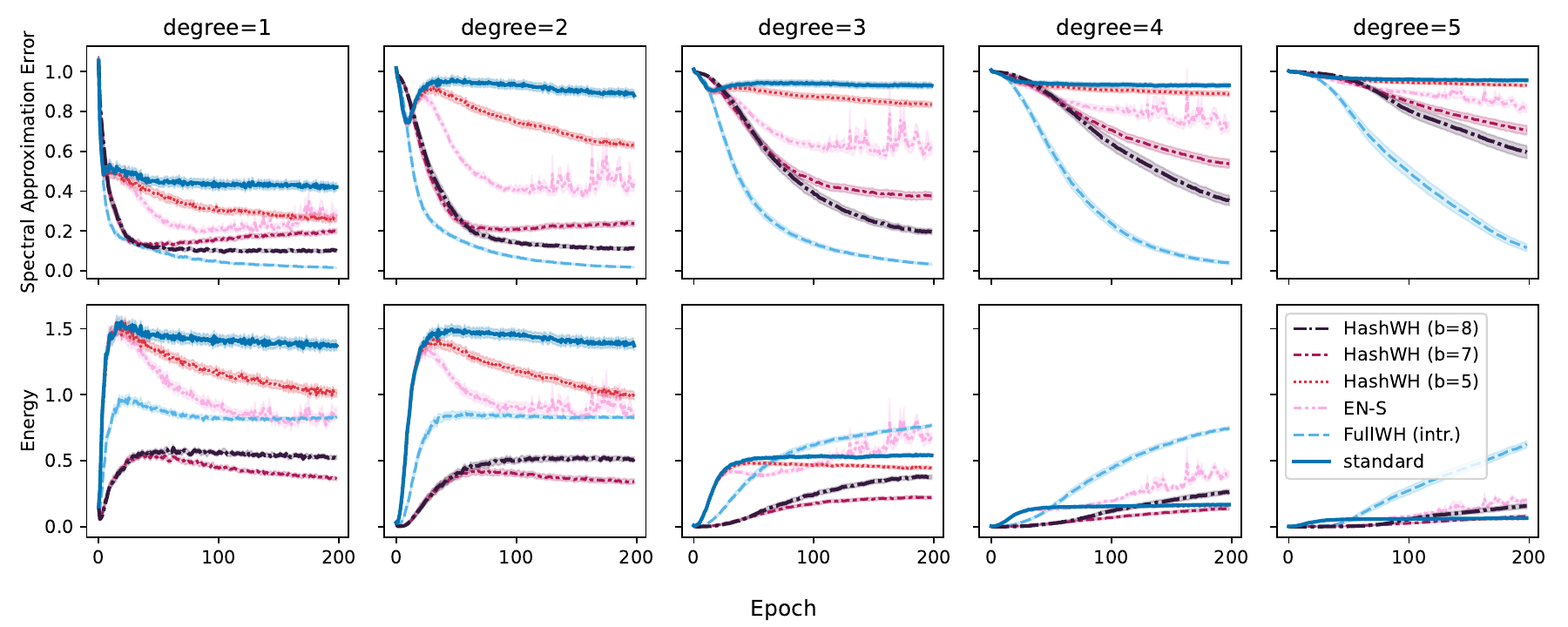}
    % \vspace*{-3mm}
  \caption{Evolution of the Spectral Approximation Error (SAE) and energy of the network during training, split by frequency degree. Firstly, in a standard neural network, the energy of high-degree frequencies is essentially zero compared to low-degree frequencies. Secondly, for low degrees (2 and 3) the energy continues to increase while the SAE exhibits overfitting behavior. This implies the neural network starts learning erroneous low-degree frequencies after some epochs. Our regularizer prevents overfitting in lower degrees and enforces higher energy on higher-degree frequencies. Regularized networks show lower energies for lower degrees and higher energy for higher degrees when compared to the standard neural network.
}
    % \vspace*{0mm}
  \label{fig:degree_split_function_error}
\end{figure*}

To gain more insight, we split the frequencies into subsets $S$ consisting of frequencies with the same degree. We visualize the evolution of SAE and also the Fourier energy of the network  defined as $\|\widehat{\mathbf{g_N}}_S\|_2^2$ in Figure~\ref{fig:degree_split_function_error}. Firstly, the energy of high-degree frequencies is essentially zero for the standard neural network when compared to the low-degree frequencies, which further substantiates the claim that  standard neural network training does not learn any high-degree frequencies. We can see that our \textsc{HashWH} regularization scheme helps the neural network learn higher degree frequencies as there is more energy in the high degree components. Secondly, looking at the lower degrees 2 and 3 we can see that the standard neural network reduces the SAE up to some point but then starts overfitting. Looking at the energy plot one can attribute the overfitting to picking up irrelevant degree 2 and 3 frequencies. We see that the regularization scheme helps prevent the neural net from overfitting on the low-degree frequencies and their SAE reduces roughly monotonously. We observe that \textsc{HashWH} (ours) with a big enough hashing matrix size exhibits the best performance among tractable methods in terms of SAE on all degrees. Finally, we can see \textsc{HashWH} is distributing the energy to where it should be for this dataset: less in the low-degree and more in the high-degree frequencies.

% To summarise, we observe that, unlike the common belief that Neural Networks are learning high-degree noises when overfitting, they tend to find a local optimum by learning simpler low-degree noises. On the contrary, the practice of inducing sparsity in the Fourier spectrum demonstrates significant potential for saving the model from this phenomenon and letting it learn relevant high-degree frequencies.
Finally, it is worth noting that our regularizer makes the neural network behave more like a \emph{decision tree}. It is well known that ensembles of decision tree models have a sparse and low-degree Fourier transform \citep{kushilevitz1991learning}. Namely, let $g:\{0,1\}^n \rightarrow \R$ be a function that can be represented as an ensemble of $T$ trees each of depth at most $d$. Then $g$ is $k=O(T \cdot 4^d)$-sparse and of degree at most $d$ (Appendix~\appref{app:subsec:tree_fourier}). Importantly, their spectrum is \emph{exactly sparse} and unlike standard neural networks, which seem to ``fill up'' the spectrum on the low-degree end, i.e., learn irrelevant low-degree coefficients, decision trees avoid this. Decision trees are well-known to be effective on discrete/tabular data \citep{tabnet}, and our regularizer prunes the spectrum of the neural network so it behaves similarly.

% \input{experiments}
% \vspace{-5mm}
\section{Experiments}\label{sec:experiments}
% \vspace{-3mm}
In this section, we first evaluate our regularization method on higher dimensional input spaces (higher $n$) on synthetically generated datasets. In this setting, \textsc{FullWH} is not applicable due to its exponential runtime in $n$. In addition, we allow varying training set sizes to showcase the efficacy of the regularizer in improving generalization at varying levels in terms of the number of training points in the dataset and especially in the low-data sample regime. Next, we move on to four real-world datasets. We first show the efficacy of our proposed regularizer \textsc{hashWH} on real-world datasets in terms of achieving better generalization errors, especially in the low-data sample regimes. Finally, using an ablation study, we experimentally convey that the low-degree bias does not result in lower generalization error.

\begin{figure*}[!htb]
    \centering
     \begin{subfigure}[b]{0.54\linewidth}
         \centering
         \includegraphics[width=\linewidth]{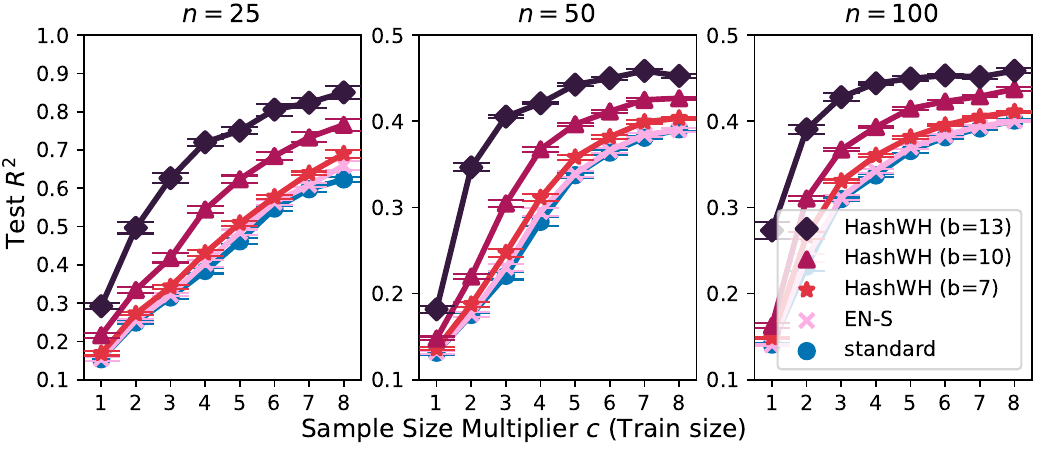}
         \caption{Generalization comparison}
         \label{fig:synthetic_large}
     \end{subfigure}
     \begin{subfigure}[b]{0.455\linewidth}
         \centering
         \includegraphics[width=\linewidth]{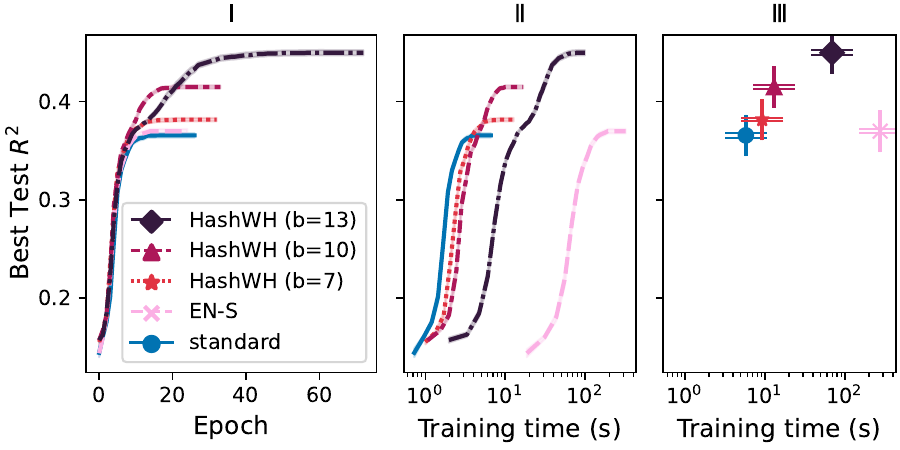}
        \caption{Runtime comparison}
         \label{fig:runtime_synthetic_large}
     \end{subfigure}
    \caption{(a) Generalization performance on learning a synthetic function $g^*:\{0,1\}^n \rightarrow \mathbb{R}$ with train set size: $c \cdot 25n$ (b) Best achievable test $R^2$ (\RomanNumeralCaps 1) at end of each epoch (\RomanNumeralCaps 2) up to a certain time (seconds). (\RomanNumeralCaps 3) Shows the early stopped $R^2$ score vs. time (seconds). We provide significant improvements across all training sizes over \textsc{EN-S} and standard neural networks, while also showing an order of magnitude speed-up compared to \textsc{EN-S}.}
\end{figure*}

\begin{figure*}[!htb]
    \centering
     \begin{subfigure}[b]{\linewidth}
         \centering
         \includegraphics[width=\linewidth]{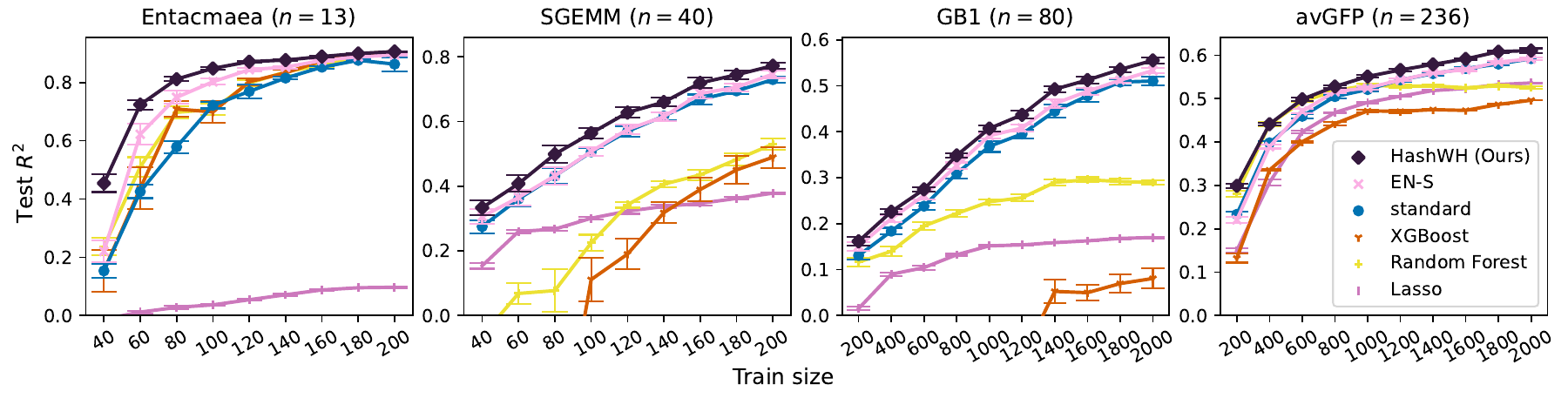}
         % \vspace*{-7mm}
         \caption{Performance on learning real datasets}
         % \vspace*{3mm}
         \hspace*{\fill}
         \label{fig:real_data_score}
     \end{subfigure}
     \begin{subfigure}[b]{0.33\linewidth}
         \centering
         \includegraphics[width=\linewidth]{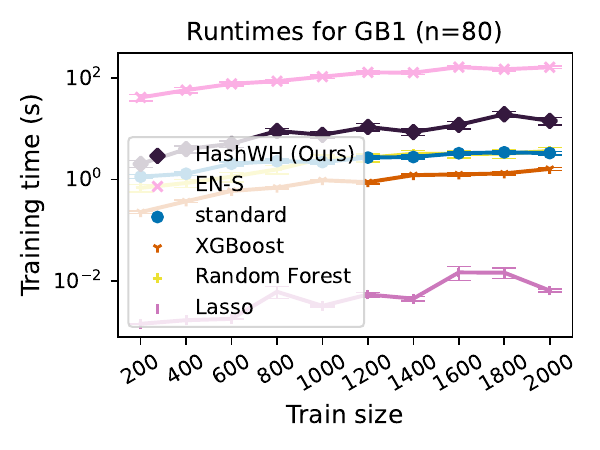}
         \caption{Runtimes for GB1}
         \label{fig:gb1_runtime}
     \end{subfigure}
         \begin{subfigure}[b]{0.41\linewidth}
         \centering
         \includegraphics[width=\linewidth]{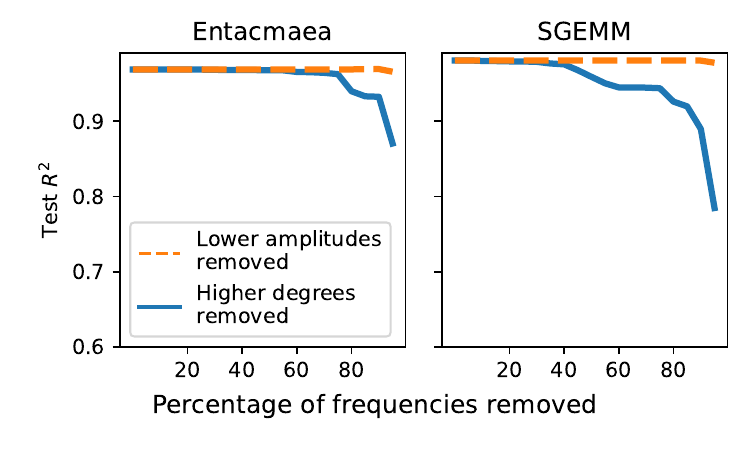}
         \caption{Ablation studies}
         \label{fig:ablation_dual}
     \end{subfigure}
    \begin{subfigure}[b]{0.25\linewidth}
         \centering
         \includegraphics[width=\linewidth]{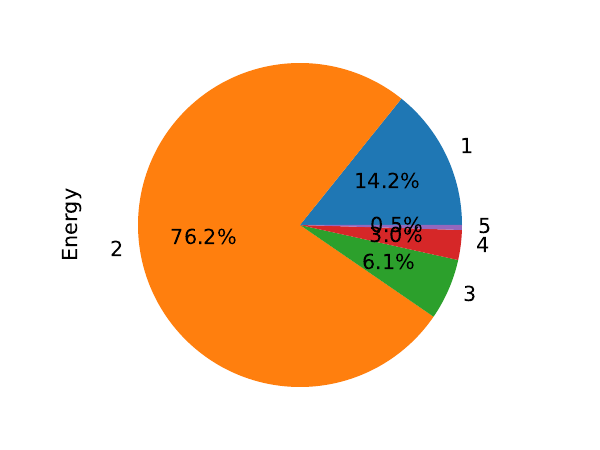}
         \caption{Energy distribution over \\degrees in Entacmaea}
         \label{fig:entacmaea_energy}
     \end{subfigure}
     % \vspace*{-4mm}
    \caption{(a) Generalization performance of standard and regularized neural networks and benchmark ML models on four real datasets. (b) Training times of different models on the GB1 dataset (c) Results of an ablation study on the potential effect of simplicity bias in the generalization error. This figure shows picking higher amplitude coefficients results in better generalization compared to picking the lower degree terms (d) Distribution of the energy over degree-based sets of frequencies in Entacmaea's top 100 Fourier coefficients. This shows high-degree components constitute a non-negligible portion of the energy of the function.
}
\label{fig:score_real_data}
\end{figure*}

% \vspace{-5mm}
\subsection{Synthetic data} 
% \vspace{-2mm}
\textbf{Setup.}
Again, we consider a synthetic pseudo-boolean target function $g^*:\{0,1\}^n \rightarrow \mathbb{R}$, which has $25$ frequencies in its support $|\text{supp}(g^*)|=25$, with the degree of maximum five, i.e., $\forall f \in \text{supp}(g^*): \text{deg}(f)\leq5$. To draw a $g^*$, we sample each of its support frequencies $f_i$ by first uniformly sampling its degree $d \sim \mathcal{U}_{\{1,2,3,4,5\}}$, based on which we then sample $f_i \sim \{f\in\{0,1\}^n|\text{deg}(f)=d \}$ and its corresponding amplitude uniformly $\widehat{g^*}(f_i) \sim \mathcal{U}_{[-1, 1]}$.

We draw $g^*$ as above for different input dimensions $n\in\{25,50,100\}$. We pick points uniformly at random from the input domain $\{0,1\}^n$ and evaluate $g^*$ to generate datasets of various sizes: we generate five independently sampled datasets of size $c \cdot 25n$, for different multipliers $c\in \{1,..,8\}$ (40 datasets for each $g^*$).
We train a 5-layer fully-connected neural network on each dataset using five different random seeds to account for the randomness in the training procedure. Therefore, for each $g^*$ and dataset size, we train and average over 25 models to capture variance arising from the dataset generation, and also the training procedure. 
%More results are in Appendix~\appref{app:subsec:synthetic_detailed}. 

\textbf{Results.} 
Figure \ref{fig:synthetic_large} shows the generalization performance of different methods in terms of their $R^2$ score on a hold-out dataset (details of dataset splits in Appendix~\appref{app:sec:technical_details}) for different dataset sizes. Our regularization method, HashWH, outperforms the standard network and \textsc{EN-S} in all possible combinations of input dimension, and dataset size. Here, EN-S does not show any significant improvements over the standard neural network, while \textsc{HashWH} (ours) improves generalization by a large margin. Moreover, its performance is tunable via the hashing matrix size $b$. 

To stress the computational scalability of \textsc{HashWH} (ours), Figure \ref{fig:runtime_synthetic_large} shows the achievable $R^2$-score by the number of training epochs and training time for different methods, when $n=50$ and $c=5$ (see Appendix~\appref{app:subsec:synthetic_detailed} for other settings). The trade-off between the training time and generalization can be directly controlled with the choice of the hashing size $b$. More importantly, comparing \textsc{HashWH} with \textsc{EN-S}, we see that for any given $R^2$ we have runtimes that are orders of magnitude smaller. This is primarily due to the very time-consuming  approximation of the Fourier transform of the network at each epoch in \textsc{EN-S}.

% \vspace{-3mm}
\subsection{Real data}
\label{sec:exp:real_data}
% \vspace{-2mm}

% (e) Distribution of energy in top 100 Fourier coefficients of Entacmaea dataset over the split of Frequencies based on degree.}
    
Next, we assess the performance of our regularization method on four different real-world datasets of varying nature and dimensionality. For baselines, we include not only standard neural networks and EN-S regularization, but also other popular machine learning methods that work well on discrete data, such as ensembles of trees. Three of our datasets are related to protein landscapes \citep{poelwijk_learning_2019,  sarkisyan_local_2016, wu_adaptation_2016} which are identical to the ones used by the proposers of \textsc{EN-S} \citep{aghazadeh_epistatic_2021}, and one is a GPU-tuning \citep{nugteren_cltune_2015} dataset. See Appendix~\appref{app:sec:datasets} for dataset details.
% We first describe the datasets used and then discuss the results.

\textbf{Results.}
Figure~\ref{fig:real_data_score} displays the generalization performance of different models in learning the four datasets mentioned, using training sets of small sizes. For each given dataset size we randomly sample the original dataset with five different random seeds to account for the randomness of the dataset sub-sampling. Next, we fit five models with different random seeds to account for the  randomness of the training procedure. One standard deviation error bars and averages are plotted accordingly over the 25 runs. It can be seen that our regularization method significantly outperforms the standard neural network as well as popular baseline methods on nearly all datasets and dataset sizes. The margin, however, is somewhat smaller than on the synthetic experiments in some cases. This may be partially explained by the distribution of energy in a real dataset (Figure \ref{fig:entacmaea_energy}), compared to the uniform distribution of energy over different degrees in our synthetic setting.

To highlight the importance of higher degree frequencies, we compute the exact Fourier spectrum of the Entacmaea dataset (which is possible, since all possible input combinations are evaluated in the dataset). 
Figure \ref{fig:entacmaea_energy} shows the energy of 100 frequencies with the highest amplitude (out of 8192 total frequencies) categorized into varying degrees. This shows that the energy of the higher degree frequencies 3 and 4 is comparable to frequencies of degree 1. However, as we showed in the previous section, the standard neural network may not be able to pick up the higher degree frequencies due to its simplicity bias (while also learning erroneous low-degree frequencies).

We also study the relationship between the low-degree spectral bias and generalization in Figure~\ref{fig:ablation_dual}. The study is conducted on the two datasets ``Entacmaea'' and ``SGEMM''. We first fit a sparse Fourier function to our training data (see Appendix~\appref{app:sec:ablation_details}). We then start deleting coefficients once according to their degree (highest to lowest and ties are broken randomly) and in another setting according to their amplitude  (lowest to highest). To assess generalization, we evaluate the $R^2$ of the resulting function on a hold-out (test) dataset. This study shows that among functions of equal complexity (in terms of size of support), functions that keep the higher amplitude frequencies as opposed to ones that keep the low-degree ones exhibit better generalization. This might seem evident according to Parseval's identity, which states that time energy and Fourier energy of a function are equal.  However, considering the fact that the dataset distribution is not necessarily uniform, there is no reason for this to hold in practice. Furthermore, it shows the importance of our regularization scheme: deviating from low-degree functions and instead aiding the neural network to learn higher amplitude coefficients \emph{regardless} of the degree. 

\textbf{Conclusion} We showed through extensive experiments how neural networks have a tendency to not learn high-degree frequencies and overfit in the low-degree part of the spectrum. We proposed a computationally efficient regularizer that aids the network in not overfitting in the low-degree frequencies and also picking up the high-degree frequencies. Finally, we exhibited significant improvements in terms of $R^2$ score on four real-world datasets compared to various popular models in the low-data regime.

% This is most probably due to the assumption of uniformity of the data distribution in our method; a typical assumption in the Compressed Sensing literature. Despite the uniform sampling done in our synthetic setting, using real data distribution can decrease the quality of the Fourier transform computed hence the gain from the regularization.

% \begin{contributions} % will be removed in pdf for initial submission,
%                       % so you can already fill it to test with the
%                       % ‘accepted’ class option
%     Briefly list author contributions.
%     This is a nice way of making clear who did what and to give proper credit.

%     H.~Q.~Bovik conceived the idea and wrote the paper.
%     Coauthor One created the code.
%     Coauthor Two created the figures.
% \end{contributions}

\begin{acknowledgements} % will be removed in pdf for initial submission,
                         % so you can already fill it to test with the
                         % ‘accepted’ class option
This research was supported in part by the NCCR Catalysis (grant number 180544), a National Centre of Competence in Research funded by the Swiss National Science Foundation.  We would also like to thank Lars Lorch and Viacheslav Borovitskiy for their detailed and valuable feedback in writing the paper.

\end{acknowledgements}

% \clearpage

\bibliography{gorji_368}
\clearpage
% \newpage
% \setlength{\belowcaptionskip}{1mm}
% \appendix
\appendix
% \input{appendix}
% % NOTE: necessary when ptmx or nomathfont class option is given
% \providecommand{\upGamma}{\Gamma}
% \providecommand{\uppi}{\pi}
\section{Walsh-Hadamard transform matrix form }
\label{app:sec:walsh_hadamard}
The Fourier analysis equation is given by:
\[\widehat{g}(f) = \frac{1}{\sqrt{2^n}} \sum\limits_{x \in \{0,1\}^n}  g(x)  (-1)^{\langle f, x \rangle}\] 
Since this transform is linear, it can be represented by matrix multiplication.  
Let $\mathbf{X}\in \{0,1\}^{2^n\times n}$ be a matrix that has the enumeration over all possible $n$-dimensional binary sequences ($\{0,1\}^n$) in some arbitrary but fixed order as its rows.  Assume $\mathbf{g}(\mathbf{X})\in \mathbb{R}^{2^n}$ to be the vector of $g$ evaluated on the rows of $\mathbf{X}$.  We can compute the Fourier spectrum as:
\[
\widehat{\mathbf{g}} = \frac{1}{\sqrt{2^n}}\mathbf{H}_n \mathbf{g}(\mathbf{X})
\] 
where $\mathbf{H_n}\in \{\pm1\}^{2^n\times 2^n}$ is an orthogonal matrix given as follows. Each row of $\mathbf{H_n}$ corresponds to some fixed frequency $f \in \{0,1\}^n$ and the elements of that row are given by $(-1)^{\langle f, x \rangle}, \forall x \in \{0,1\}^n$, where the ordering of the $x$ is the same as the fixed order used in the rows of $\mathbf{X}$. The ordering of the rows in $\mathbf{H}_n$, i.e. the ordering of the frequencies considered, is arbitrary and determines the order of the Fourier coefficients in the Fourier spectrum $\widehat{\mathbf{g}}$.

It is common to define the Hadamard matrix $\mathbf{H}_n \in \{\pm1\}^{2^n\times 2^n}$ through the following recursion:
\begin{equation*}
    \mathbf{H}_n = \mathbf{H}_2 \otimes\mathbf{H}_{n-1},
\end{equation*}
where $\mathbf{H}_2 := \begin{bmatrix}
  1 & 1\\ 
  1 & -1
\end{bmatrix}$, and $\otimes$ is the Kronecker product. We use this in our implementation. This definition corresponds to the ordering similar to $n$-bit binary numbers (e.g., $[0,0,0], [0,0,1], [0,1,0], ..., [1,1,1]$ for $n=3$) for both frequencies and time (input domain). 

Computing the Fourier spectrum of a network using a matrix multiplication lets us utilize a GPU and efficiently compute the transform, and its gradient and conveniently apply the back-propagation algorithm. 
% \section{HashWH Details}
\section{Algorithm Details}
\label{app:sec:HashWH_details}

Let $g:\{0,1\}^n \rightarrow \mathbb{R}$ be a pseudo-boolean function with Fourier transform $\widehat{g}$. In the context of our work, this pseudo-boolean function is the neural network function. One can sort the Fourier coefficient of $g$ according to magnitude, from biggest to smallest, and consider the top $k$ biggest coefficients as the most important coefficients. This is because they capture the most energy in the Fourier domain and by Parseval's identity also in the time (original input) domain. It is important to us that these $k$ coefficients $\widehat{g}(f_1), \dots, \widehat{g}(f_k)$ are not hashed into the same bucket. Say for example two large coefficients $\widehat{g}(f_i), \widehat{g}(f_j), i \neq j$ end up in the same bucket, an event which we call a \emph{collision}. If they have different signs, their sum can form a cancellation and the $L_1$ norm will enforce their sum to be zero. This entails an approximation error in the neural network: Our goal is to sparsify the Fourier spectrum of the neural network and ``zero out'' the non-important (small-magnitude) coefficients, not to impose wrong constraints on the important (large magnitude) coefficients. 

With this in mind, we first prove our hashing result Equation~\ref{eq:hash_sum} from Section~\ref{subsec:wht_background}. Next, we provide guarantees on how increasing the hashing bucket size reduces collisions. Furthermore, we show how independently sampling the hashing matrix over different rounds guarantees that each coefficient does not collide too often. Ideas presented there can also be found in \cite{alon1999linear, amrollahi_efficiently_2019}. We finally review \textsc{EN-S} and showcase the superiority and scalability of our method in terms of computation. 

\subsection{Proof of Equation~\ref{eq:hash_sum}}
\label{app:subsec:hash_sum_proof}
Let 
\[u_\mathbf{\sigma}(\tilde{x}) = \sqrt{\frac{2^n}{2^b}} g(\mathbf{\sigma} \tilde{x}), \forall \tilde{x} \in \{0,1\}^b\]
as in Section~\ref{subsec:wht_background}.

We can compute its Fourier transform $\widehat{u}_\mathbf{\sigma}(\Tilde{f})$  as:
\begin{align}
    \widehat{u}_\mathbf{\sigma}(\Tilde{f}) &= \frac{1}{\sqrt{2^b}} \sum_{\tilde{x} \in \{0, 1\}^b} u_\mathbf{\sigma}(\tilde{x})(-1)^{\langle \Tilde{f}, \tilde{x} \rangle} \nonumber\\
    &= \frac{1}{\sqrt{2^b}} \sum_{\tilde{x} \in \{0, 1\}^b}  \sqrt{\frac{2^n}{2^b}} \  g(\mathbf{\sigma} \tilde{x})(-1)^{\langle \Tilde{f}, \tilde{x} \rangle} \nonumber\\
    &= \frac{\sqrt{2^n}}{2^b} \sum_{\tilde{x} \in \{0, 1\}^b}  g(\mathbf{\sigma} \tilde{x})(-1)^{\langle \Tilde{f}, \tilde{x} \rangle} \label{eq:hash_sum_expansion}
\end{align}

Inserting the Fourier expansion of $g$ into Equation~\eqref{eq:hash_sum_expansion} we have:
\begin{align*}
    \widehat{u}_\mathbf{\sigma}(\Tilde{f}) &= \frac{1}{2^b} \sum_{\tilde{x} \in \{0, 1\}^b}  (-1)^{\langle \Tilde{f}, \tilde{x} \rangle}\sum_{f \in \{0,1\}^n}\widehat{g}(f)(-1)^{\langle f, \mathbf{\sigma} \tilde{x} \rangle} \nonumber\\
    &= \frac{1}{2^b} \sum_{\tilde{x} \in \{0, 1\}^b}  \sum_{f \in \{0,1\}^n} \widehat{g}(f)(-1)^{\langle \mathbf{\sigma}^\top f, \tilde{x} \rangle} (-1)^{\langle \Tilde{f}, \tilde{x} \rangle}  \nonumber\\
    & = \frac{1}{2^b} \sum_{f \in \{0,1\}^n} \widehat{g}(f) \sum_{\tilde{x} \in \{0, 1\}^b}  (-1)^{\langle \mathbf{\sigma}^\top f + \Tilde{f}, \tilde{x} \rangle}
\end{align*}
The second summation is always zero unless $\mathbf{\sigma}^\top f + \Tilde{f}=0$, i.e., $\mathbf{\sigma}^\top f=\Tilde{f}$, in which case the summation is equal to $2^b$. Therefore:
\begin{align*}
    \widehat{u}_\mathbf{\sigma}(f) = \sum_{\tilde f \in \{0,1\}^n:\ \mathbf{\sigma}^T \tilde f=f} \widehat{g}(f)
\end{align*}

\subsection{Collisions for \textsc{HashWH}}
\label{app:subsec:collision_probability}
We first review the notion of \emph{pairwise independent} families of hash functions introduced by \cite{carter1979universal}. We compute the expectation of the number of collisions for this family of hash functions. We then show that uniformly sampling $\sigma \in \{0,1\}^{n\times b}$ in our hashing procedure (in \textsc{HashWH}) gives rise to a pairwise independent hashing scheme. 

\begin{definition}[Pairwise independent hashing]\label{def:pairwise_independent}
Let $\mathcal{H} \subseteq \{h| h \in \{0,1\}^n\rightarrow\{0,1\}^b\}$ be a family of hash functions.  Each hash function maps $n$-dimensional inputs $x\in\{0,1\}^n$ into a $b$-dimensional buckets $u=h(f)\in\{0,1\}^b$ and is picked uniformly at random from $\mathcal{H}$. We call this family \emph{pairwise independent} if for any distinct pair of inputs $f_1 \neq f_2\in\{0,1\}^n$ and an arbitrary pair of buckets $u_1, u_2\in\{0,1\}^b$:
\begin{enumerate}
    \item $P(h(f_1)=u_1)=\frac{1}{2^b}$
    \item $P((h(f_1)=u_1) \land (h(f_2)=u_2))=\frac{1}{2^{2b}}$
\end{enumerate}
\end{definition}
(randomness is over the sampling of the hash function from $\mathcal{H}$)

Assume $S=\{f_1, ..., f_k\} \subseteq \{0,1\}^n$ is a set of $k$ arbitrary elements to be hashed using the hash function $h \in \{0,1\}^n\rightarrow\{0,1\}^b$ which is sampled from a pairwise independent hashing family. Let $c_{ij}$ be an indicator random variable for the collision of $f_i, f_j, i \neq j$, i.e., $c_{ij} = \begin{cases}
      1 & h(f_i)=h(f_j) \\
      0 & h(f_i) \neq h(f_j)
    \end{cases}\,$, for $i \neq j \in [k]$.

\begin{lemma}
\label{lemma:collision_count}
The expectation of the total number of collisions $C=\sum_{i \neq j \in [k]} c_{ij}$ in a pairwise independent hashing scheme is given by: 
$\mathbb{E}[C]=\frac{(k-1)^2}{2^{b}}$.
\end{lemma}
\begin{proof}
\begin{align*}
    \mathbb{E}[C]&=\sum_{i \neq j \in [k]} \mathbb{E}[c_{ij}]\\
    &= \sum_{i \neq j \in [k]} \sum_{u \in \{0,1\}^b} P((h(f_i)=u) \land (h(f_j)=u)) \\
    &= \frac{(k-1)^2}{2^b},
\end{align*}
where we have applied the linearity of expectation. 
\end{proof}
The next Lemma shows that the hashing scheme of \textsc{HashWH} introduced in Section~\ref{subsec:Hashwh} is also a pairwise independent hashing scheme. However, there is one small caveat: the hash function always maps $0 \in \{0,1\}^n$ to $0 \in \{0,1\}^b$ which violates property 1 of the pairwise independence Definition \ref{def:pairwise_independent}. If we remove $0$ from the domain then it becomes a pairwise independent hashing scheme. 
\begin{lemma}
\label{lemma:HashWH_pairwise}
The hash function used in the hashing procedure of our method \textsc{HashWH}, i.e., $h(.)=\mathbf{\sigma}^\top(.)$ where $\mathbf{\sigma} \sim \mathcal{U}_{\{0,1\}^{n\times b}}$ is a hashing matrix whose elements are sampled independently and  uniformly at random (with probability $\frac{1}{2}$)  from $\{0,1\}$, is pairwise independent if we exclude $f=0$ from the domain.
\end{lemma}
\begin{proof}
Note that for any input $f\in\{0,1\}^n, f \neq 0$, its hash $\mathbf{\sigma}^\top f$ is a linear combination of columns of $\mathbf{\sigma}^\top$, where $f$ determines the columns. We denote $i$\textsuperscript{th} column of $\mathbf{\sigma}^\top$ by $\mathbf{\sigma}^\top_{\bullet i}$.
Let $f$ be non-zero in $t\geq 1$ positions (bits) $\{i_1, ..., i_t\}\subseteq[n]$. The value of $h(f)$ is equal to the summation of the columns of $\mathbf{\sigma}^\top$ that corresponds to those $t$ positions: $\mathbf{\sigma}^\top_{\bullet i_1}, \cdots, \mathbf{\sigma}^\top_{\bullet i_t}$. Let $u \in \{0,1\}^b$ be an arbitrary bucket. The probability the sum of the columns equals $u$ is $\frac{1}{2^b}$ as all sums are equally likely i.e. 
\[P(h(x) = u) = \frac{1}{2^b}\]
    
Let $f_1, f_2 \neq 0, f_1 \neq f_2 \in \{0,1\}^n$ be a pair of distinct non-zero inputs. Since $f_1$ and $f_2$ differ in at least one position (bit), $h(f_1)$ and $h(f_2)$ are independent random variables. Therefore, for any arbitrary $u_1, u_2 \in \{0,1\}^b$
\begin{align*}
&P(h(f_1)=u_1 \land h(f_2)=u_2) \\
= &P(h(f_1)=u_1)P(h(f_2)=u_2) = \frac{1}{2^{2b}}
\end{align*}
\end{proof}

Lemmas \ref{lemma:collision_count} and \ref{lemma:HashWH_pairwise} imply that the expected total number of collisions $C$ in hashing frequencies of the top $k$ coefficients of $g$ in our hashing scheme is also equal to: $\mathbb{E}[C] = \frac{(k-1)^2}{2^b}$. Our guarantee shows that the number of collisions goes down linearly in the number of buckets $2^b$. 

Finally, let $f_1$ be an important frequency i.e. one with a large magnitude $|\widehat{g}(f_1)|$. By independently sampling a new hashing matrix $\mathbf{\sigma}$ at each round of back-prop, we avoid always hashing this frequency into the same bucket as some other important frequency. By a union bound on the pairwise independence 
property, the probability that a frequency $f_1$ collides with any other frequency $f_2, \dots, f_k$ is upper bounded by $\frac{k-1}{2^b}$. Therefore, over $T$ rounds of back-prop the number of times this frequency collides follows a binomial distribution with $p \leq \frac{k-1}{2^b}$ ($\frac{k-1}{2^b}<1$ for a large enough $b$). We denote the number of times frequency $f_1$ collides over the $T$ rounds as $C_{f_1}$. The expected number of collisions is $\mu \triangleq Tp$ which goes down linearly in the number of buckets. With a Chernoff bound we can say that roughly speaking, the number of collisions we expect can not be too much larger than a fraction $p$ of the $T$ rounds.

By a Chernoff's bound we have:
\[
P(C_{f_1} \geq (1+\delta )\mu ) \leq e^{-\frac{\delta^{2}\mu}{2+\delta}}
\]
where $\mu=Tp$ as mentioned before 

For examples setting $\delta =1$
\[
P(C_{f_1} \geq 2 \mu ) \leq e^{-\frac{\mu}{3}}
\]
As $T \rightarrow \infty$ this probability goes to zero. This means that the probability that the number of times the frequency collides during the $T$ rounds to not be more than a fraction $(1+\delta)p=2p$ of the time is, for all practical purposes, essentially zero. Building on this intuition, we can see that for any fixed $0<\epsilon<1$, setting $b=\log_2(\frac{k-1}{\epsilon})$ guarantees that collision of a given frequency happens on average a fraction $\epsilon$ of the $T$ rounds and not much more. 
% For each bucket $f_u \in \{0, 1\}^b$ with only one frequency $f \in \text{supp}(g_\theta)$ hashed into, $\widehat{u}_\mathbf{\sigma}(f_u)=\widehat{g_\theta}(f)$. Therefore, we can approximate the non-zero part of $\widehat{\mathbf{g_\theta}}$ using $\widehat{\mathbf{u_\mathbf{\sigma}}}$, where the lower number of collisions leads to a better approximation. In case of no collisions, $\|\widehat{\mathbf{u_\mathbf{\sigma}}}\|_1=\|\widehat{\mathbf{g_\theta}}\|_1$, which has the probability of $P(C=0)\geq 1-\frac{k^2}{2^b}$ (Markov's inequality).
\subsection{\textsc{EN-S} details}
\label{app:subsec:EN-S}
To avoid computing the exact Fourier spectrum of the network at each back-propagation iteration in \textsc{FullWH}, \cite{aghazadeh_epistatic_2021} suggest an iterative regularization technique to enforce sparsity in the Fourier spectrum of the network called \textsc{EN-S}. 

We first briefly describe the Alternating Direction Method of Multipliers \citep{boyd_distributed_2011} (ADMM) which is an algorithm that is used to solve convex optimization problems. This algorithm is used to derive \textsc{EN-S}. Finally, we discuss \textsc{EN-S} itself and highlight the advantages of using our method \textsc{HashWH} over it.

\paragraph{ADMM.} Consider the following separable optimization objective:
\begin{align*}
    \min_{\bm{x}\in\mathbb{R}^n,\bm{z}\in\mathbb{R}^m} f(\bm{x}) + g(\bm{z}) \\ \text{subject to  } \mathbf{A}\bm{x}+\mathbf{B}\bm{z}=\mathbf{c},
\end{align*}
where $\mathbf{A}\in\mathbb{R}^{p\times n}$, $\mathbf{B}\in\mathbb{R}^{p\times m}$, $\mathbf{c}\in\mathbb{R}^{p}$, and $f\in \mathbb{R}^m\rightarrow \mathbb{R}$ and $g\in \mathbb{R}^n\rightarrow \mathbb{R}$ are arbitrary \emph{convex} functions 
%(w.l.o.g., $\mathbb{R}$ can be replaced by any set of interest in the domains). 
The augmented Lagrangian of this objective is formed as:
\begin{align*}
L_\rho(\bm{x}, \bm{z}, \mathbf{\gamma})=f(\bm{x})+g(\bm{z})+\mathbf{\gamma}^\top (\mathbf{A}\bm{x}+\mathbf{B}\bm{z}-\mathbf{c}) \\+ \frac{\rho}{2} \|\mathbf{A}\bm{x}+\mathbf{B}\bm{z}-\mathbf{c}\|_2^2,
\end{align*}
where $\mathbf{\gamma}\in\mathbb{R}^{p}$ are the dual variables. 

Alternating Direction Method of Multipliers \citep{boyd_distributed_2011}, or in short \emph{ADMM}, optimizes the augmented Lagrangian by alternatively minimizing it over the two variables $\bm{x}$ and $\bm{z}$ and applying a dual variable update:
\begin{align}
    \bm{x}^{k+1} &= \operatorname*{argmin}_{\bm{x}\in\mathbb{R}^n} L_\rho(\bm{x}, \bm{z}^k, \mathbf{\gamma}^k) & (\bm{x}\text{-minimization})\nonumber\\
    \bm{z}^{k+1} &= \operatorname*{argmin}_{\bm{z}\in\mathbb{R}^m} L_\rho(\bm{x}^{k+1}, \bm{z}, \mathbf{\gamma}^k) & (\bm{z}\text{-minimization})\nonumber\\
    \mathbf{\gamma}^{k+1} &= \mathbf{\gamma}^{k} + \rho(\mathbf{A}\bm{x}^{k+1}+\mathbf{B}\bm{z}^{k+1}-\mathbf{c})  & (\text{dual var. update}) \nonumber
\end{align}

In a slightly different formulation of ADMM, known as \emph{scaled-dual} ADMM, the dual variable can be scaled which results in a similar optimization scheme:
\begin{align}
    \bm{x}^{k+1} &= \operatorname*{argmin}_{\bm{x}\in\mathbb{R}^n} f(\bm{x}) + \frac{\rho}{2}\|\mathbf{A}\bm{x} + \mathbf{B}\bm{z}^k-\mathbf{c}+\mathbf{\gamma}^k \|^2_2 \nonumber\\
    \bm{z}^{k+1} &= \operatorname*{argmin}_{\bm{z}\in\mathbb{R}^m} g(\bm{z}) + \frac{\rho}{2}\|\mathbf{A}\bm{x}^{k+1} + \mathbf{B}\bm{z}-\mathbf{c}+\mathbf{\gamma}^k \|^2_2 \nonumber\\
    \mathbf{\gamma}^{k+1} &= \mathbf{\gamma}^{k} + \mathbf{A}\bm{x}^{k+1}+\mathbf{B}\bm{z}^{k+1}-\mathbf{c}\label{eq:scaled_dual_ADMM}
\end{align}

Using ADMM, one can decouple the joint optimization of two separable groups of parameters into two alternating separate optimizations for each individual group.

\paragraph{\textsc{EN-S}.}
To apply ADMM, \cite{aghazadeh_epistatic_2021} reformulate the \textsc{FullWH} loss, by introducing a new variable $\bm{z}$ and adding a constraint such that it is equal to the Fourier spectrum:
\begin{align*}
&\mathcal{L}_{EN-S}=\mathcal{L}_{net}+\lambda\|\bm{z}\|_1 \\
&\text{subject to:   } \bm{z}=\widehat{\mathbf{g_\theta}}=\mathbf{H}_n\mathbf{g_\theta}(\mathbf{X})
\end{align*}
,  where $g_{\bm{\theta}}$ is the neural network parameterized by $\bm{\theta}$, $\mathbf{H}_n\in\{0,1\}^{2^n\times2^n}$ is the Hadamard matrix, and $\mathbf{X}\in\{0,1\}^{2^n\times n}$ is the matrix of the enumeration over all points on the Boolean cube $\{0,1\}^n$. 

They use the scaled-dual ADMM~\eqref{eq:scaled_dual_ADMM} followed by a few further adjustments to reach the following alternating scheme for optimization of $\mathcal{L}_{EN-S}$:
\begin{align}
    \bm{\theta}^{k+1} &= \operatorname*{argmin}_{\bm{\theta}} \mathcal{L}_{net} + \frac{\rho}{2}\|\mathbf{g_\theta}(\mathbf{X}_T) - \mathbf{H}_T\bm{z}^k+\mathbf{\gamma}^k \|^2_2 \nonumber\\
    \bm{z}^{k+1} &= \operatorname*{argmin}_{\bm{z}} \lambda \|\bm{z}\|_0 + \frac{\rho}{2}\|\mathbf{g}_{\bm{\theta}^{k+1}}(\mathbf{X}_T) - \mathbf{H}_T\bm{z}+\mathbf{\gamma}^k \|^2_2 \nonumber\\
    \mathbf{\gamma}^{k+1} &= \mathbf{\gamma}^{k} + \mathbf{g}_{\bm{\theta}^{k+1}}(\mathbf{X}_T) - \mathbf{H}_T\bm{z}^{k+1}, \label{eq:scaled_dual_ENS}
\end{align}
where $\mathbf{X}_T\in \{0,1\}^{O(2^mn)\times n}$ is the input enumeration matrix $\mathbf{X}\in\{0,1\}^{2^n\times n}$ sub-sampled at $O(2^mn)$ rows, $\mathbf{H}_T\in \{0,1\}^{O(2^mn)\times n}$ is the Hadamard matrix $\mathbf{H}_n\in\{0,1\}^{2^n\times 2^n}$ subsampled at similar $O(2^mn)$ rows, and $\mathbf{\gamma}\in\mathbb{R}^{O(2^mn)}$ is the dual variable. We will introduce the hash size parameter $m$ momentarily.

Using the optimization scheme \eqref{eq:scaled_dual_ENS}, they decouple the optimization of $\mathcal{L}_{EN-S}$ into two separate alternating optimizations: 1) minimizing $\mathcal{L}_{net}$ by fixing $\bm{z}$ and optimizing network parameters using SGD for an epoch ($\bm{\theta}$-minimization), 2) fixing $\theta$ and computing a sparse Fourier spectrum approximation of the network at the end of each epoch and updating the dual variable ($\bm{z}$-minimization).

To approximate the sparse Fourier spectrum of the network at $\bm{z}$-minimization step, they use the ``SPRIGHT'' algorithm \citep{li_spright_2015}. SPRIGHT requires $O(2^mn)$ samples from the network to approximate its Fourier spectrum and runs with the complexity of $O(2^mn^3)$, where $m$ is the hash size used in the algorithm (the equivalent of $b$ in our setting).
In \textsc{EN-S} optimization scheme \eqref{eq:scaled_dual_ENS}, these $O(2^mn)$ inputs are denoted by the matrix $\mathbf{X}_T\in \{0,1\}^{O(2^mn)\times n}$, and are fixed during the whole optimization process. This requires the computation of the network output on these $O(2^mn)$ inputs at each back-prop iteration in $\bm{\theta}$-minimization, as well as at the end of each epoch to run SPRIGHT in $\bm{z}$-minimization.

\paragraph{\textsc{EN-S} vs. \textsc{HashWH}.}
The hashing done in our method, \textsc{HashWH}, is basically the first step of many (if not all) sparse Walsh-Hadamard transform approximation methods \citep{li2015active,scheibler2015fast,li_spright_2015,amrollahi_efficiently_2019}, including \textsc{SPRIGHT} \citep{li_spright_2015} that is used in \textsc{EN-S}. In the task of sparse Fourier spectrum approximation, further, extra steps are done to infer the \emph{exact} frequencies of the support and their associated Fourier coefficients. These steps are usually computationally expensive. Here, since we are only interested in the $L1$-norm of the Fourier spectrum of the network and are not necessarily interested in retrieving the exact frequencies in its support, we found the idea of approximating it with the $L1$-norm of the Fourier spectrum of our hash function compelling. This is the core idea behind \textsc{HashWH} which lets us stick to the \textsc{FullWH} formulation using a scalable approximation of the $L1$-norm of the network's Fourier spectrum.

From the mere computational cost perspective, \textsc{EN-S} requires a rather expensive sparse Fourier spectrum approximation of the network at the end of each epoch. We realized, one bottleneck of their algorithm was the evaluation of the neural network on the required time samples of their sparse Fourier approximation algorithm. We re-implemented this part on a GPU to make it substantially faster. Still, we empirically observe that more than half of the run time of each \textsc{EN-S} epoch is spent on the Fourier transform approximation. Furthermore, in \textsc{EN-S}, the network output needs to be computed for $\Omega(2^mn)$ samples at each back-prop iteration. 

On the contrary, in \textsc{HashWH}, the network Fourier transform approximation is not needed anymore. We only compute the network output on precisely $2^b$ samples at each round of back-propagation to compute the Fourier spectrum of our sub-sampled neural network. Remember that our $b$ is roughly equivalent to their $m$. Since the very first step in their sparse Fourier approximation step is a hashing step into $2^m$ buckets. 

Let us compare our method with \textsc{EN-S} more concretely. For the sake of simplicity, we ignore the network sparse Fourier approximation step ($\bm{z}$-minimization) that happens at the end of each epoch for \textsc{EN-S} and assume their computational complexity is only dominated by the $\Omega(2^m n)$ evaluations made during back-prop. In order to use the same number of samples as \textsc{EN-S}, we can set our hashing size to $b=m+log(n)+c$, where $c$ is a constant which we found in practice to be at least $c\geq 3$. In the case of our avGFP experiment, this would be for instance $b\geq18$ in \textsc{HashWH} for \textsc{EN-S} with $m=7$. There, we outperformed \textsc{EN-S} using $b\in\{7, 10, 13, 16\}$ in terms of $R^2$-score. Note that even with $b=18$ we are still at least two times faster than \textsc{EN-S} as we do not go the extra mile of approximating the Fourier spectrum of the network at each epoch. 

%Finally, we allow for a new hashing matrix at the beginning of each round of back-prop. The \textsc{SPRIGHT} uses a deterministic hash matrix which is not   Therefore we are more robust to collisions also are not constrained to use a fixed sample set to approximate the Fourier spectrum of the network, which makes \textsc{HashWH} statistically more robust against ``harmful'' collisions in the hashing procedure (see Section~\ref{app:subsec:collision_probability}).

% Furthermore, some of ADMM assumptions, for instance, the separability of the parameter groups considered as well as the convexity of the functions and linearity of the constraint with respect to the parameters, do not seem to be met in the formulation of \textsc{EN-S}, as it originally stands. Therefore, we believe it requires more theoretical investigations before guaranteeing the expected results obtained through ADMM.

\section{Datasets}
\label{app:sec:datasets}
We list all the datasets used in the real dataset Section~\ref{sec:exp:real_data}. 
\paragraph{Entacmaea quadricolor fluorescent protein. (Entacmaea)}
\cite{poelwijk_learning_2019} study the fluorescence brightness of all $2^{13}$ distinct variants of the Entacmaea quadricolor fluorescent protein, mutated at $13$ different sites. 
They examine the goodness of fit ($R^2$-score) when only using a limited set of frequencies of the highest amplitude. They report that only $1\%$ of the frequencies are enough to describe data with a high goodness of fit ($R^2=0.96$), among which multiple high-degree frequencies exist.
% They report the existence of multiple high-degree frequencies in its support (referred to as high-order epistatic interactions), given the observed data.

\paragraph{GPU kernel performance (SGEMM).}
\cite{nugteren_cltune_2015} measures the running time of a matrix product using a parameterizable SGEMM GPU kernel, configured with different parameter combinations. The input has $14$ categorical features that we one-hot encode into $40$-dimensional binary vectors.

\paragraph{Immunoglobulin-binding domain of protein G (GB1).}
\cite{wu_adaptation_2016} study the ``fitness'' of variants of protein GB1, that are mutated at four different sites. Fitness, in this work, is a quantitative measure of the stability and functionality of a protein variant. Given the $20$ possible amino acids at each site, they report the fitness for $20^4=160,000$ possible variants, which we represent with one-hot encoded 80-dimensional binary vectors. In a noise reduction step, they included $149,361$ data points as is and replaced the rest with imputed fitness values. We use the former, the untouched portion, for our study.

\paragraph{Green fluorescent protein from Aequorea victoria (avGFP).}
\cite{sarkisyan_local_2016} estimate the fluorescence brightness of random mutations over the green fluorescent protein sequence of Aequorea victoria (avGFP) at 236 amino acid sites. We transform the data into the boolean space of the absence or presence of a mutation at each amino acid site by averaging the brightness for the mutations with similar binary representations. This converts the original $54,024$ distinct amino acid mutations into $49,089$ 236-dimensional binary data points.

\section{Implementation technical details}
\label{app:sec:technical_details}
\paragraph{Neural network architecture and training}
We used a 5-layer fully connected neural network including both weights and biases and LeakyReLu as activations in all settings. For training, we used MSE loss as the loss of the network in all settings. We always initialized the networks with Xavier uniform distribution.  We fixed 5 random seeds in order to make sure the initialization was the same over different settings. The Adam optimizer with a learning rate of $0.01$ was used for training all models. We always used a single Nvidia GeForce RTX 3090 to train each model to be able to fairly compare the runtime of different methods. 
%We use early stopping for picking the model using which we report the test $R^2$ (details in the following data split segment).
% We run all synthetic experiments done for Fourier spectrum evolution experiments for 500 epochs (we reported up to 300 epochs in the figures for the sake of clearer visualization)
We did not utilize other regularization techniques such as Batch Normalization or Dropout to limit our studies to analyze the mere effect of Fourier spectrum sparsification. We use networks of different widths in different experiments which we detail in the following:
\begin{itemize}
    \item \textit{Fourier spectrum evolution:} The architecture of the network is $10 \times 100 \times 100 \times 10 \times 1$.
    \item \textit{High-dimensional synthetic data:} For each $n \in \{25, 50, 100\}$, the architecture of the network is $n \times 2n \times 2n \times n \times 1$.
    \item \textit{Real data:} Assuming $n$ to be the dimensionality of the input space, we used the network architecture of $n \times 10n \times 10n \times n \times 1$ for all the experiments except avGFP. For avGFP with $n=236$, we had to down-size the network to $n \times n \times n \times n \times 1$ to be able to run \textsc{EN-S} on GPU as it requires a significant amount of samples to compute the Fourier transform at each epoch in this dimension scale.
\end{itemize}

\paragraph{Data splits}
In the Fourier spectrum evolution experiment, where we do not report $R^2$ of the predictions, we split the data into training and validation sets (used for hyperparameter tuning). For the rest of the experiments, we split the data into three splits training, validation, and test sets. We use the validation set for the hyperparameter tuning (mainly the regularizer multiplier $\lambda$ and details to be explained later) and early stopping. We stop each training after 10 consecutive epochs without any improvements over the best validation loss achieved and use the epoch with the lowest loss for testing the model. All the $R^2$s reported are the performance of the model on the (hold-out) test set. 

% We always continued the experiments until the stop due to no improvement happens.

For each experiment, we used different training dataset sizes that are explicitly mentioned in the main body of the paper. Here we list the validation and test dataset sizes:
\begin{itemize}
    \item \textit{Fourier spectrum evolution:} Given that $n=10$ and the Boolean cube is of size $2^n=1024$, we always use the whole data and split it into training and validation sets. For example, for the training set of size $200$, we use the rest of the $824$ data points as the validation set.
    \item \textit{High-dimensional synthetic data:} For each training set, we use validation and test sets of five times the size of the training set. That is, for a training set of size $c \cdot 25n$, both of our validation and test sets are of size $c \cdot 125n$.
    \item \textit{Real data:} After taking out the training points from the dataset, we split the remaining points into two sets of equal sizes one for validation and one for test. 
    %Given $2^{13}=8192$ data points in the Entacmaea dataset, we divide the data not used in the training set into two equal-size sets of validation and test for this dataset. For example for the training set of size $200$, validation and test sets would be of size $3996$. For the rest of the datasets, we use validation and test sets of size $20,000$.
\end{itemize}

\paragraph{Hyper-parameter tuning.}
In all experiments, we hand-picked candidates for important hyper-parameters of each method studied and tested every combination of them, and picked the version with the best performance on the validation set.
This includes testing different $\lambda \in \{0.0001, 0.001, 0.01, 0.1\}$ for HashWH, $\lambda \in \{0.01, 0.1, 1\}$ and $\rho \in \{0.001, 0.01, 0.1\}$ for \textsc{EN-S}, and $\lambda \in \{0.01, 0.1, 1\}$ for \textsc{FullWH}. Furthermore, we also used the following hyper-parameters for the individual experiments:
\begin{itemize}
    \item \textit{Fourier spectrum evolution:} We used $b\in\{5, 7, 8\}$ for HashWH and $m=5$ for the \textsc{EN-S}. We did not tune $b$ for HashWH as we reported all the results in order to show the graceful dependence with increasing the hashing matrix size. 
    \item \textit{High-dimensional synthetic data:} We used $b\in\{7, 10, 13\}$ for HashWH and $m=7$ for the \textsc{EN-S}. We did not tune $b$ for HashWH as we reported each individually.
    \item \textit{Real data:} We used $b\in\{7, 10, 13\}$ for HashWH and $m=7$ for \textsc{EN-S} in the Entacmaea, SGEMM, and GB1 experiments. Furthermore, for avGFP, we also considered $b=16$ for HashWH. Unlike the synthetic experiments, where we reported results for each $b$ individually, we treated $b$ as a hyper-parameter in real data experiments. For Lasso, we tested different L1 norm coefficients of $\lambda \in \{10^{-5}, 10^{-4}, 10^{-3}, 10^{-2}, 10^{-1}, 1\}$. For Random Forest, we tested different numbers of estimators $n_{estimators} \in \{100, 200, 500, 1000 \}$, and different maximum depths of estimators $max_{depth} \in \{5, 10, 15\}$ for Entacmaea experiments and $max_{depth} \in \{10, 20, 30, 40, 50\}$ for the rest of experiments. We tested the exact same hyper-parameter candidates we considered for Random Forest in our XGBoost models.
\end{itemize}
Like common practice, we always picked the hyper-parameter combination resulting in the minimum loss on the validation set, and reported the model's performance on the test (hold-out) dataset.

%\paragraph{Plots.}
%In the plots representing the Fourier spectrum (e.g. Figures \labelcref{fig:data_freq_heatmap,fig:sc_heatmap}), the spectrums are averaged over multiple runs ($125$ runs for Figure \ref{fig:data_freq_heatmap} and $25$ runs for Figure \ref{fig:sc_heatmap}). For the rest of the plots reporting Test $R^2$ or runtimes, the values are averaged over multiple runs for each specific setting and are accompanied by error bars representing the standard error intervals.

\paragraph{Code repositories.}
All the implementations for the methods as well as the experiments are publicly accessible through \href{https://github.com/agorji/WHRegularizer}{https://github.com/agorji/WHRegularizer}. 
% All the implementations for our methods as well as the experiments will be publicly accessible in the final version of the paper after the double-blind review process is over, by making the GitHub repository of the project publicly available.

For \textsc{EN-S} and \textsc{FullWH} regularizers, we used the implementation shared by \cite{aghazadeh_epistatic_2021}\footnote{https://github.com/amirmohan/epistatic-net}. We applied minor changes so to compute samples needed for the Fourier transform approximation in \textsc{EN-S} on GPU, making it run faster and fairer to compare our method with.

We used the python implementation of \verb|scikit-learn|\footnote{https://scikit-learn.org} for our Lasso and Random Forest experiments. We also used the XGBoost\footnote{https://xgboost.readthedocs.io} python library for our XGBoost experiments.

% \begin{itemize}
%     \item hadamard lambdas used for different methods and experiments
%     \item learning rate and optimizer
%     \item scaling of synthetic dataset
% \end{itemize}

\section{Ablation study details}
\label{app:sec:ablation_details}
To study the effect of the low-degree simplicity bias on generalization on the real-data distribution, we conduct an ablation study by fitting a sparse Fourier transform to two of our datasets. To this end, we fit Random Forest models on Entecamaa and SGEMM datasets, such that they achieve test $R^2$ of nearly 1 on an independent test set not used in the training. Then, we compute the exact sparse Fourier transform of each Random Forest model, which essentially results in a sparse Fourier function that has been fitted to the training dataset. In our ablation study, finally, we remove frequencies based on two distinct regimes of lower-amplitudes-first and higher-degrees-first and show that the former harms the generalization more. This is against the assumption of simplicity bias being always helpful.

In the next two subsections, we provide the details on how to compute the exact sparse Fourier transform of a Random Forest model as well as finer details of the study setup.

\subsection{Fourier transform of (ensembles) of decision trees}\
\label{app:subsec:tree_fourier}
% In the previous section, we saw how the Fourier representation provides us with insight in the form of Shapley values, helps to identify \emph{complementary} and \emph{substitutable} features, and how {\em Fourier sparsity} helps for their {\em efficient calculation}. In this section, we go even further and review how decision trees, 
% %convey why it makes sense to compile these representations into decision trees, 
% a class of models celebrated for their interpretability, naturally admit sparse Fourier representations \citep{kushilevitz1993learning}. 

%. This approach is motivated by the fact that that decision trees admit sparse Fourier representations \citep{kushilevitz1993learning}. Therefore, one can expect the compiled decision trees would accurately approximate the Fourier transform and hence the original function. We continue by providing more context on decision trees.
A decision tree, in our context, is a rooted binary tree whose nodes can have either zero or two children. Each leaf node is assigned a real number. Each non-leaf node corresponds to one of $n$ binary features. The tree defines a function $t:\{0, 1\}^n \rightarrow  \mathbb{R}$ in the following way: To compute $t(x)$ we look at the root,  corresponding to, say, feature $i \in [n]$. Next, we check the value of the variable $x_i$. If the value of the variable is equal to $0$ we look at the left child. If it is equal to $1$ we look at the right child. Then we repeat this process until we reach a leaf. The value of the function $t$ evaluated at $x$ is the real number assigned to that leaf. In all that follows, when referring to decision trees, we will denote them by the function $t:  \{0, 1\}^n \rightarrow  \mathbb{R}$. 

Given a decision tree, we can compute its Fourier transform recursively. Let $i \in [n]$ denote the feature corresponding to its root. Then the tree can be represented as follows:
\begin{equation} 
\label{eq:recursive_tree}
t(x) = \frac{1+(-1)^{\langle e_i, x \rangle}}{2} t_{\mathrm{left}}(x) + \frac{1-(-1)^{\langle e_i, x \rangle}}{2} t_{\mathrm{right}}(x)
\end{equation}

Hereby, $t_{\mathrm{left}}:\{0,1\}^{n-1}\rightarrow \R$ and $t_{\mathrm{right}}:\{0,1\}^{n-1} \rightarrow \R$ are the left and right sub-trees respectively. Therefore, one can recursively compute the Fourier transform of a decision tree. 
% or even an ensemble of trees (e.g., Random Forest) very efficiently. 

This also portrays why a decision tree of depth $d$ is a function of degree $d$. Moreover, for each tree $t$, if $|\text{supp}(t_{left})|=k_{\mathrm{left}}$ and $|\text{supp}(t_{right})|=k_{\mathrm{right}}$, then $|\text{supp}(t)| \leq 2(k_{\mathrm{left}}+k_{\mathrm{right}})$. This implies that a decision tree is $k$-sparse with $k=O(4^d)$. However, in many cases, when the decision tree is not complete or cancellations occur, the Fourier transform is even sparser.

Finally, we can also compute the Fourier transform of an ensemble of trees such as one produced by the random forest and XGBoost algorithms. In the case of regression, the ensemble just predicts the average prediction of its constituent trees.  Therefore its Fourier transform is the (normalized) sum of the Fourier transforms of its trees as well. If a random forest model consists of $T$ different trees then its Fourier transform is $k=O(T 4^d)$-sparse and of degree equal to its maximum depth. 
% This often occurs, e.g., in the case of classification  where labels (leaf values) are zero and one. Nevertheless, in all cases $k=O(4^d)$, which is polynomial in the size of the tree, since a tree of depth $d$ contains at most $2^d$ nodes.     

\subsection{Ablation study setup}
For the Entacmae dataset, we used a training set of size $5,000$ and a test set of size $2,000$, for which we trained a Random Forest model with $100$ trees with maximum depths of $7$. For the SGEMM dataset, we used a training set of size $100,000$ and a test set of size $5,000$, for which we trained a Random Forest model with $100$ trees with a maximum depth of $10$.

\section{Extended experiment results}
\label{app:extended_results}
% Uncomment figures last to avoid reducing the compilation speed ...
%In the main body of the article, for each plot, we usually reported one version out of all variations of a similar experiment that are different in traits such as the train size used or the version of the random synthetic function considered. 
Here, we report the extended experiment results containing variations not reported in the main body of the paper.

\subsection{Fourier spectrum evolution}
\label{app:subsec:evolution_detailed}
\begin{figure*}[h]
    \centering
    \begin{subfigure}[b]{0.75\linewidth}
         \centering
         \includegraphics[width=\linewidth]{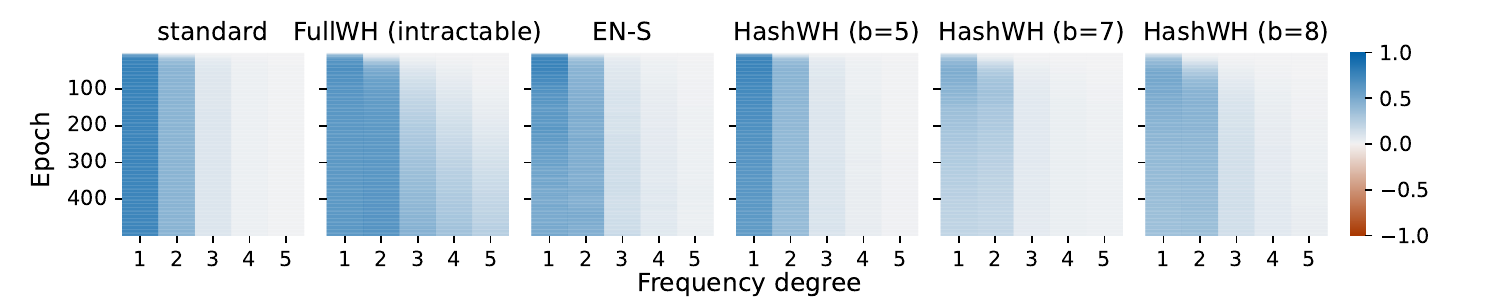}
         \caption{$\text{Train size}=100$}
     \end{subfigure}
    \begin{subfigure}[b]{0.75\linewidth}
         \centering
         \includegraphics[width=\linewidth]{plots/spectrum/data_freq_heatmap_n10_d5_size2.pdf}
         \caption{$\text{Train size}=200$}
     \end{subfigure}
    \begin{subfigure}[b]{0.75\linewidth}
         \centering
         \includegraphics[width=\linewidth]{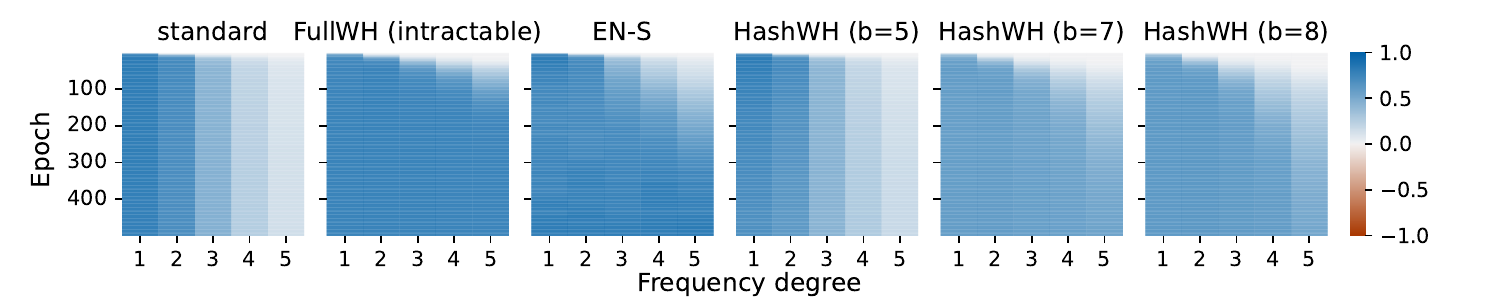}
         \caption{$\text{Train size}=300$}
     \end{subfigure}
    \begin{subfigure}[b]{0.75\linewidth}
         \centering
         \includegraphics[width=\linewidth]{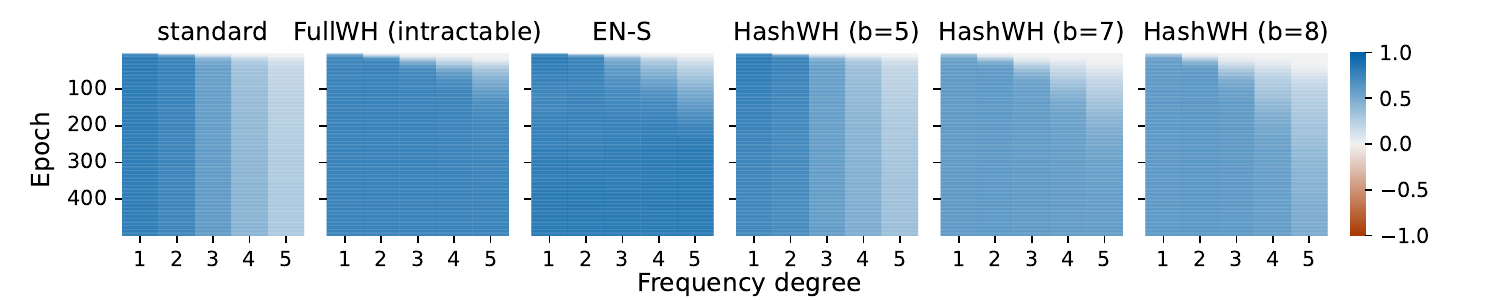}
         \caption{$\text{Train size}=400$}
     \end{subfigure}
     \caption{Evolution of the Fourier spectrum during training limited to the target support, using training sets of different sizes. All synthetic functions have single frequencies of each degree in their support that are all given the amplitude of $1$. This is an extended version of Figure~\ref{fig:data_freq_heatmap}, where we only reported the results for the train set size $200$. It can be observed that the Fourier-sparsity-inducing (regularized) methods are \emph{always} better than the standard neural network in picking up the higher-degree frequencies, regardless of the training size. Each method shows better performance when trained on a larger training set in terms of converging at earlier epochs and also converging to the true Fourier amplitude it is supposed to.}
     \label{fig:full_data_freq_heatmap}

\end{figure*}

We randomly generated five synthetic target functions $g^*\in\{0,1\}^{10}$ of degree $d=5$, each having a single frequency of each degree in its support (the randomness is over the choice of support). We create a dataset by randomly sampling the Boolean cube. 
Figure~\ref{fig:full_data_freq_heatmap} shows the evolution of the Fourier spectrum of the learned neural network function for different methods over training on datasets of multiple sizes ($100, 200, 300, 400$) limited to the target support. This is the extended version of Figure~\ref{fig:data_freq_heatmap}, where we only reported results for the train size of $200$. We observe that, quite unsurprisingly, each method shows better performance when trained on a larger training set in terms of converging at earlier epochs and also converging to the true Fourier amplitude it is supposed to. It can also be observed that the Fourier-sparsity-inducing (regularized) methods are \emph{always} better than the standard neural network in picking up the higher-degree frequencies, regardless of the training size.

\begin{figure*}[h]
    \centering
\begin{subfigure}[b]{0.8\linewidth}
         \centering
         \includegraphics[width=\linewidth]{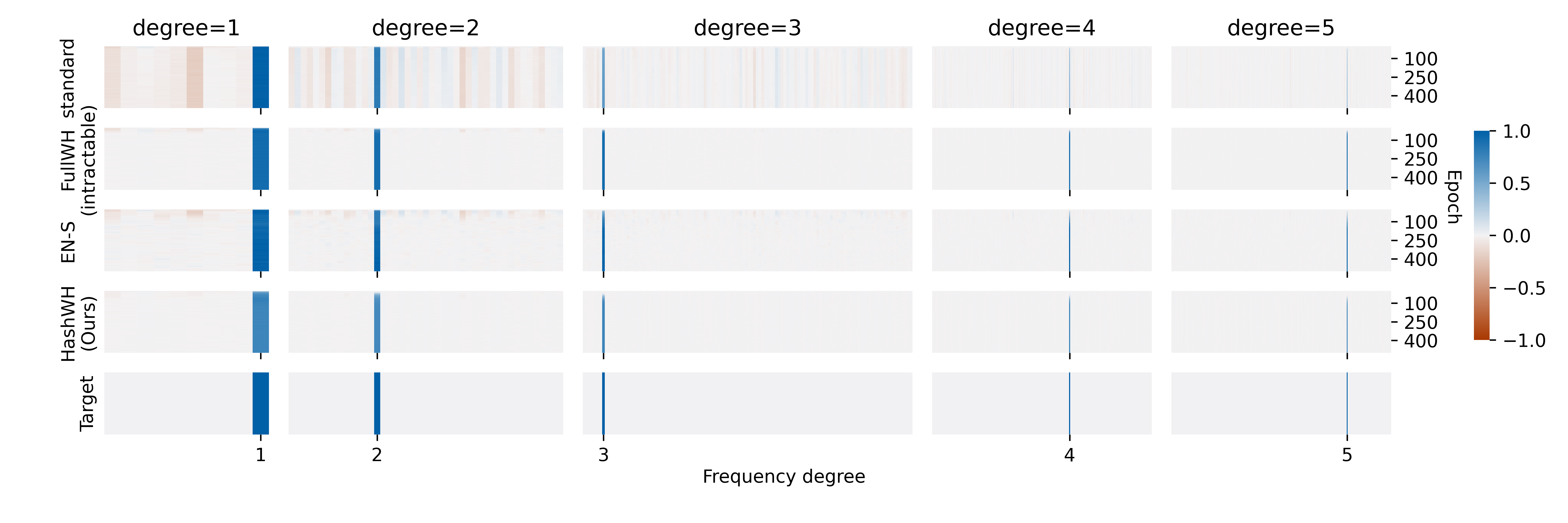}
     \end{subfigure}
\begin{subfigure}[b]{0.8\linewidth}
         \centering
         \includegraphics[width=\linewidth]{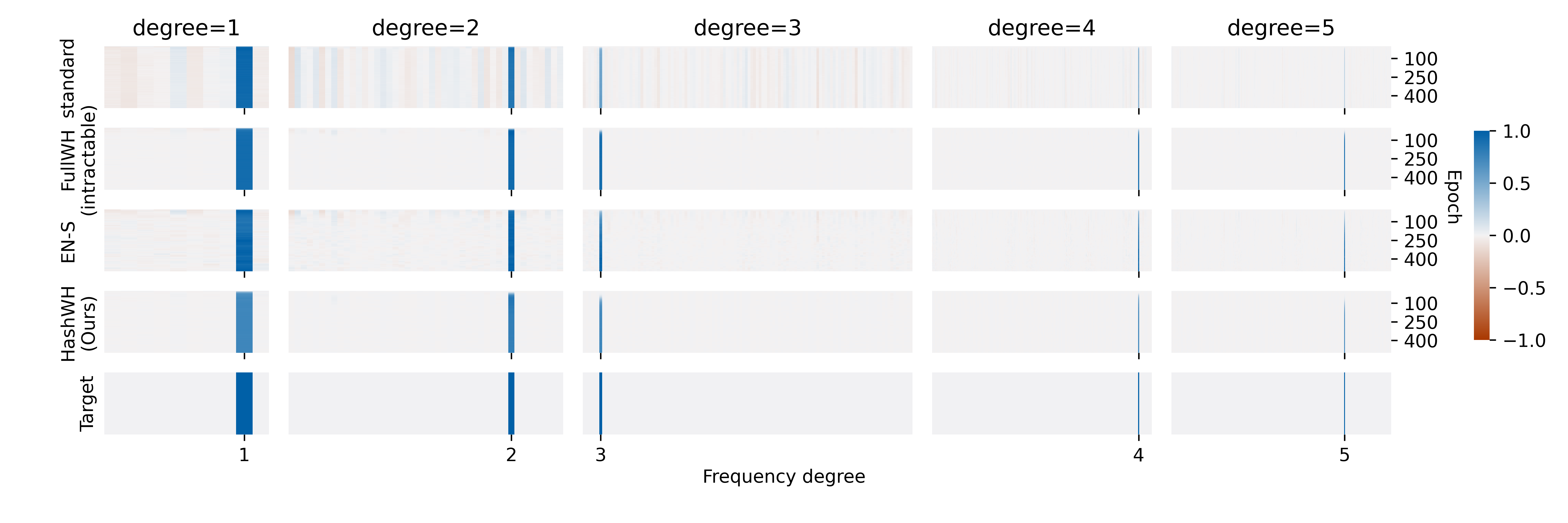}
     \end{subfigure}
\begin{subfigure}[b]{0.8\linewidth}
         \centering
         \includegraphics[width=\linewidth]{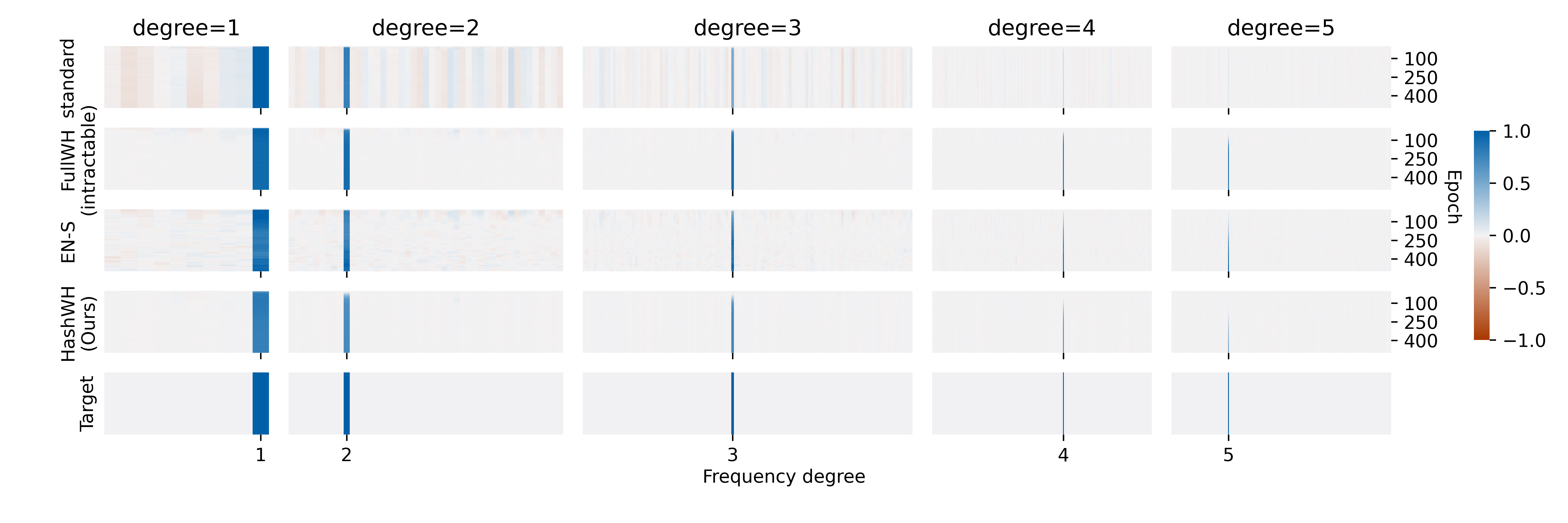}
     \end{subfigure}
\begin{subfigure}[b]{0.8\linewidth}
         \centering
         \includegraphics[width=\linewidth]{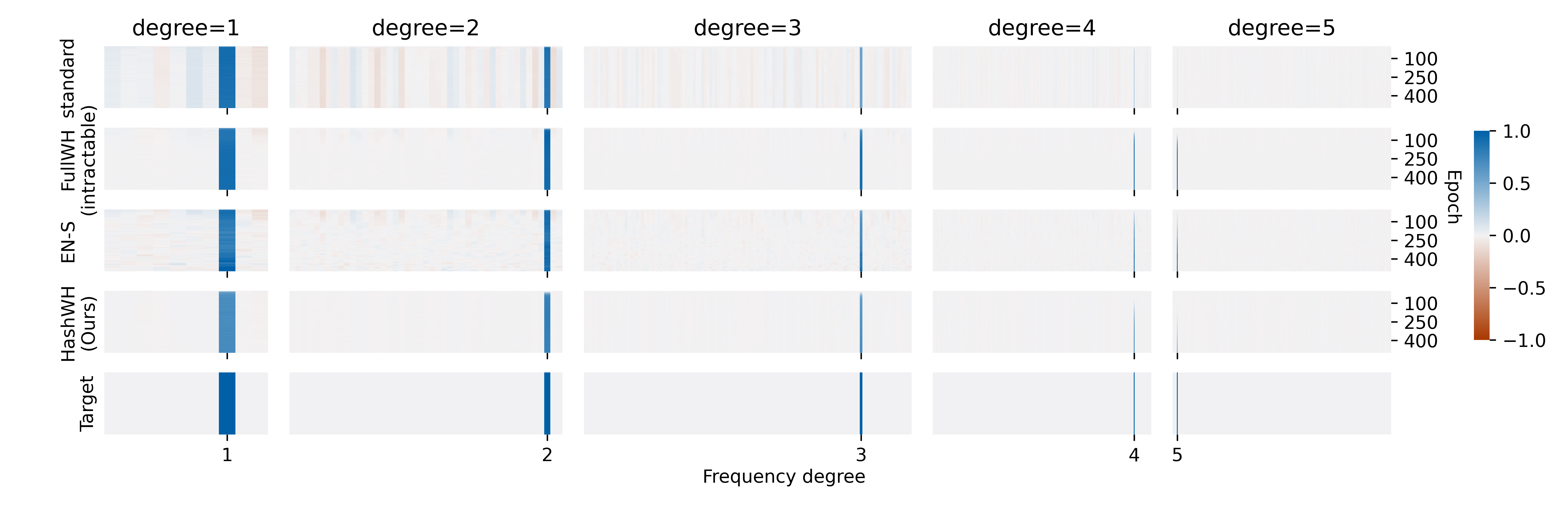}
     \end{subfigure}
     \caption{Evolution of the Fourier spectrum in learning a synthetic function $g^*\in\{0,1\}^{10}$ of degree $5$ during training, categorized by frequency degree. All synthetic functions used have single frequencies of each degree in their support that are all given the amplitude of $1$. We reported the results for one draw of $g^*$ in Figure \ref{fig:sc_heatmap} and the four others here, for the training dataset size of $200$. In addition to the incapability of the standard neural network in learning high-degree frequencies, they tend to consistently pick up wrong low-degree frequencies. Both of the problems are remedied through our regularizer.}
     \label{fig:full_sc_heatmap}
\end{figure*}

Figure~\ref{fig:full_sc_heatmap} goes a step further and shows the evolution of the full Fourier spectrum (not just the target frequencies) over the course of training. Here, unlike the previous isolated setting where we were able to aggregate the results from different target functions (because of always having a single frequency of each degree in the support), we have to separate the results for each target function $g^*\in\{0,1\}^{10}$, as each has a unique set of frequencies in its support.
In Figure~\ref{fig:data_freq_heatmap}, we reported the results for one version of the target function $g^*$ and Figure~\ref{fig:full_sc_heatmap} shows the Fourier spectrum evolution for the other four. We observe that in addition to the spotted inability of the standard neural network in learning higher-degree frequencies, it seems to start picking up erroneous low-degree frequencies as well.

\begin{figure*}[h]
  \centering
      \begin{subfigure}[b]{0.49\linewidth}
         \centering
            \includegraphics[width=\linewidth]{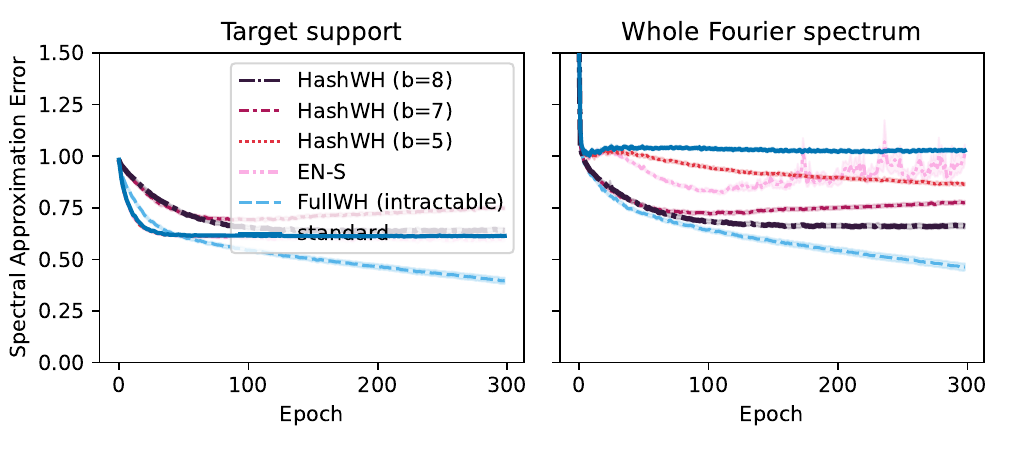}
         \caption{$\text{Train size}=100$}
     \end{subfigure}
      \begin{subfigure}[b]{0.49\linewidth}
         \centering
            \includegraphics[width=\linewidth]{plots/spectrum/function_error_n10_d5_size2.pdf}
         \caption{$\text{Train size}=200$}
     \end{subfigure}
      \begin{subfigure}[b]{0.49\linewidth}
         \centering
            \includegraphics[width=\linewidth]{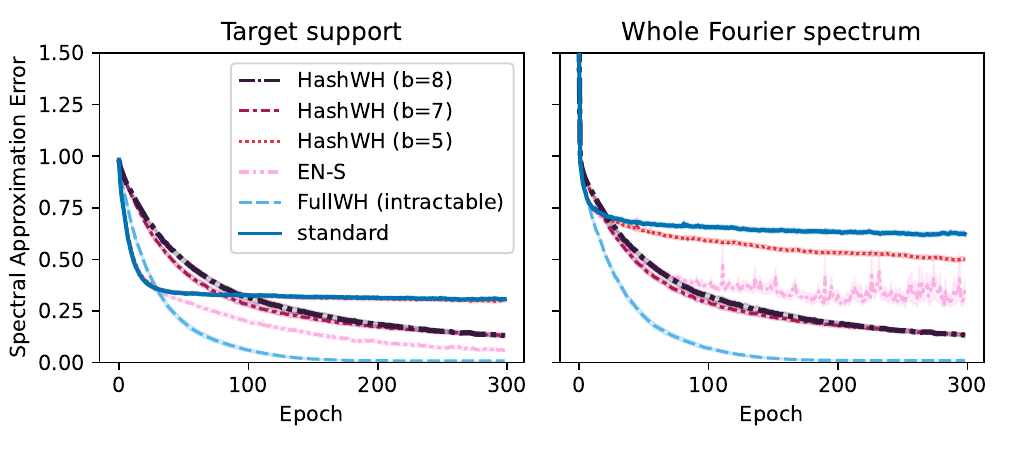}
         \caption{$\text{Train size}=300$}
     \end{subfigure}
      \begin{subfigure}[b]{0.49\linewidth}
         \centering
            \includegraphics[width=\linewidth]{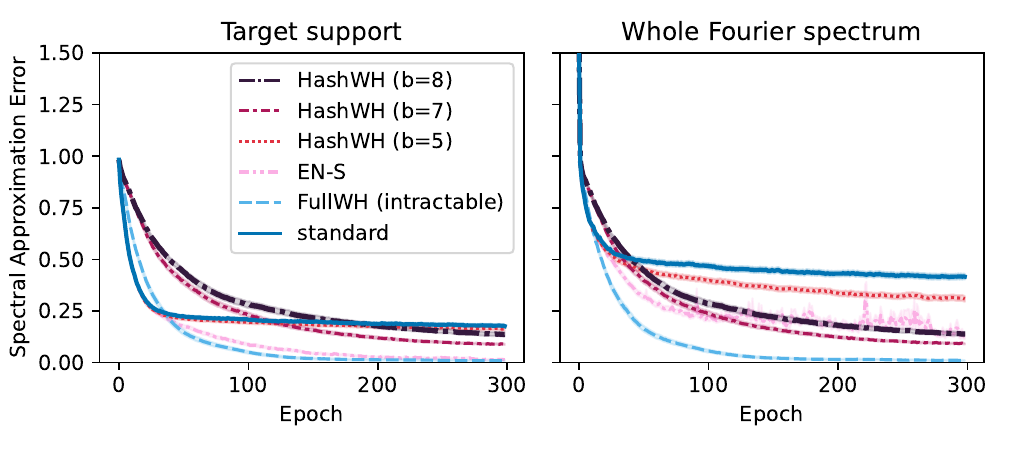}
         \caption{$\text{Train size}=400$}
     \end{subfigure}
     \caption{Evolution of the spectral approximation error (SAE) during training. The left plot limits the error to the target support, while the right one considers the whole Fourier spectrum. This is an extended version of Figure \ref{fig:function_error}, where we only reported results for train size $200$. The standard neural network is able to achieve a lower (better) (train size $100$) or somewhat similar (train size $400$) SAE on the \emph{target support} compared to our method. However, our method always achieves lower SAE on the \emph{whole Fourier spectrum}, regardless of $b$ used. This shows how our regularisation method is effective in preventing the network from learning the wrong frequencies that are not in the support.}
  \label{fig:full_function_error}
\end{figure*}

To quantitatively validate our findings, in Figure~\ref{fig:full_function_error}, we show the evolution of Spectral Approximation Error (SAE) during training on both target support and the whole Fourier spectrum. This is an extended version of Figure~\ref{fig:function_error}, where we report the results for the train size of $200$. Here we also include results when using training datasets of three other train sizes $\{100, 300, 400\}$. We observe that even though the standard neural network exhibits comparable performance to \textsc{HashWH} on the target support when the training dataset size is $100$ and $400$, it is always underperforming \textsc{HashWH} when broadening our view to the whole Fourier spectrum, regardless of the train size and the hashing size. 

From a more fine-grained perspective, in Figure~\ref{fig:full_degree_split_function_error}, we categorize the frequencies into subsets of the same degree and show the evolution of SAE and energy on each individual degree. This is an extended version of Figure~\ref{fig:degree_split_function_error}, where we reported the results for the training dataset size of $200$. Firstly, we observe that using more data aids the standard neural network to eventually put more energy on higher-degree frequencies. But it is still incapable of appropriately learning higher-degree frequencies. Fourier-sparsity inducing methods, including ours, show significantly higher energy in the higher degrees. Secondly, No matter the train size, we note that the SAE on low-degree frequencies first decreases and then increases and the standard neural network starts to overfit. This validates our previous conclusion that the standard neural network learns erroneous low-degree frequencies. Our regularizer prevents overfitting in lower degrees. Its performance of which can be scaled using the hashing size parameter $b$.

\subsection{High-dimensional synthetic data}
\label{app:subsec:synthetic_detailed}
\begin{figure*}[h]
    \centering
     \begin{subfigure}[b]{0.33\linewidth}
         \centering
         \includegraphics[width=\linewidth]{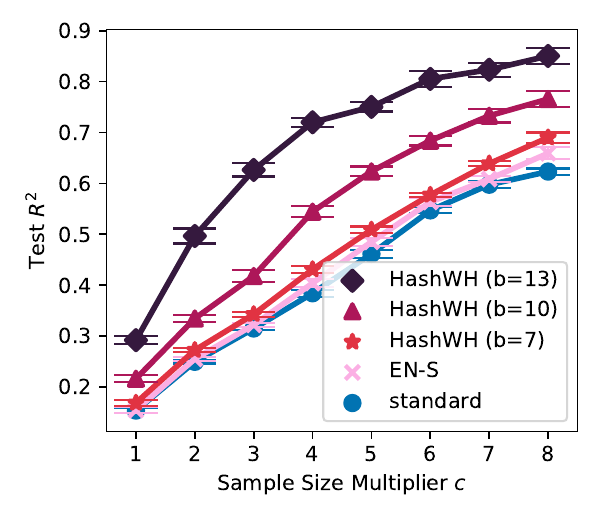}
         \caption{$n=25$, first draw of $g^*$}
     \end{subfigure}
     \begin{subfigure}[b]{0.33\linewidth}
         \centering
         \includegraphics[width=\linewidth]{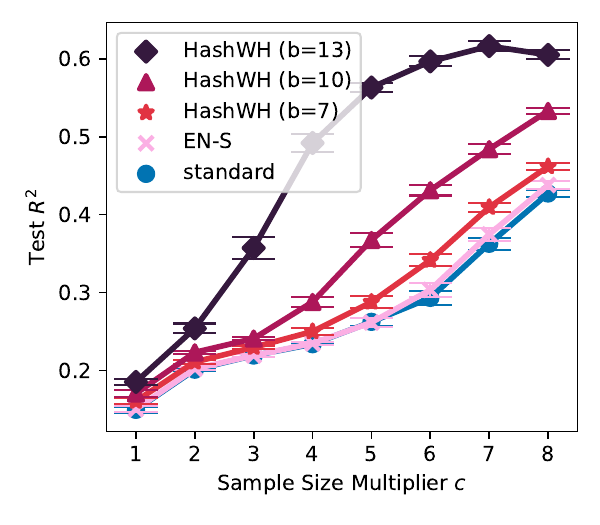}
         \caption{$n=25$, second draw of $g^*$}
     \end{subfigure}
    \begin{subfigure}[b]{0.33\linewidth}
         \centering
         \includegraphics[width=\linewidth]{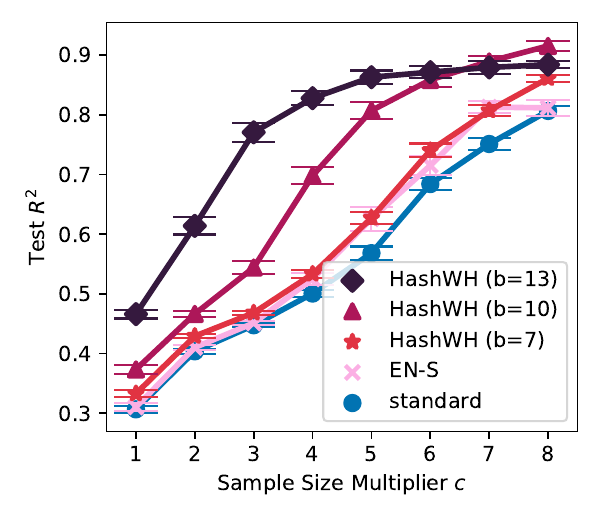}
         \caption{$n=25$, third draw of $g^*$}
     \end{subfigure}
     \begin{subfigure}[b]{0.33\linewidth}
         \centering
         \includegraphics[width=\linewidth]{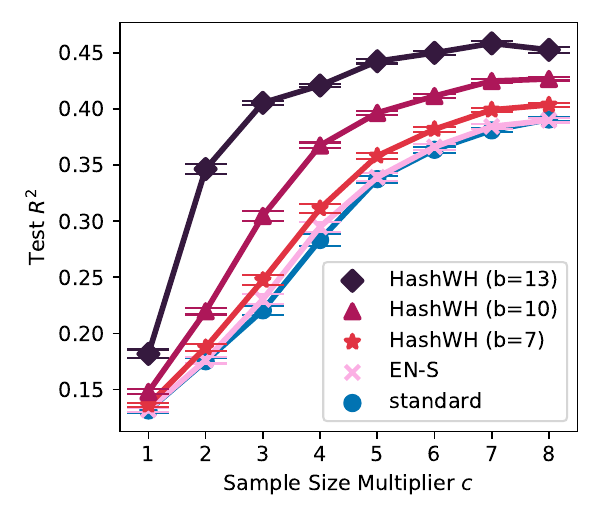}
         \caption{$n=50$, first draw of $g^*$}
     \end{subfigure}
     \begin{subfigure}[b]{0.33\linewidth}
         \centering
         \includegraphics[width=\linewidth]{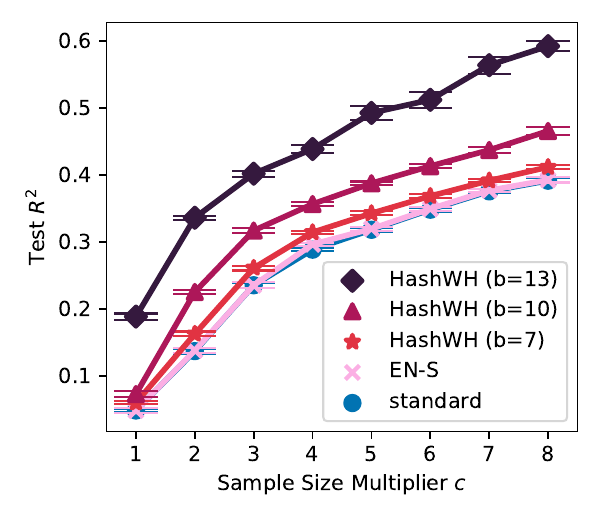}
         \caption{$n=50$, second draw of $g^*$}
     \end{subfigure}
    \begin{subfigure}[b]{0.33\linewidth}
         \centering
         \includegraphics[width=\linewidth]{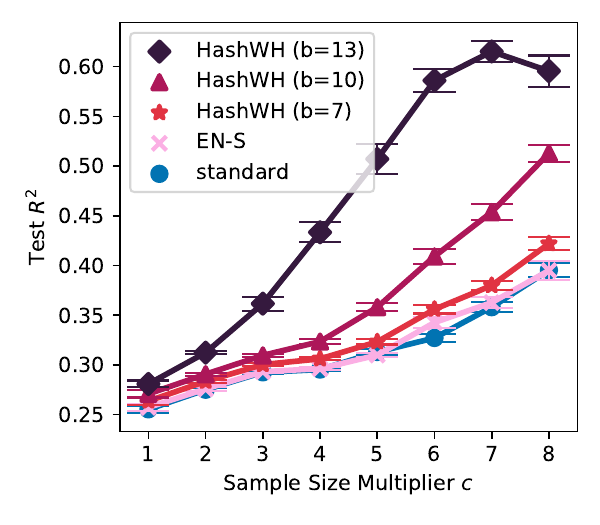}
         \caption{$n=50$, third draw of $g^*$}
     \end{subfigure}
    \begin{subfigure}[b]{0.33\linewidth}
         \centering
         \includegraphics[width=\linewidth]{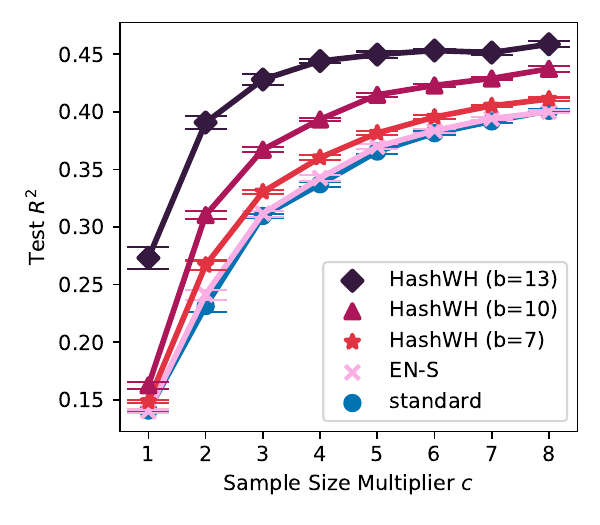}
         \caption{$n=100$, first draw of $g^*$}
     \end{subfigure}
     \begin{subfigure}[b]{0.33\linewidth}
         \centering
         \includegraphics[width=\linewidth]{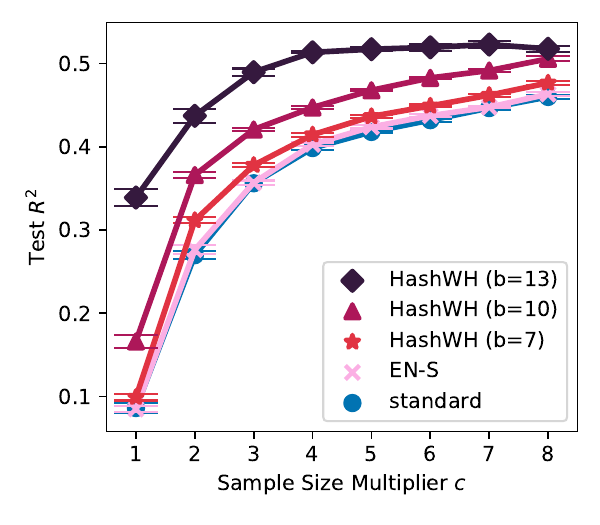}
         \caption{$n=100$, second draw of $g^*$}
     \end{subfigure}
    \begin{subfigure}[b]{0.33\linewidth}
         \centering
         \includegraphics[width=\linewidth]{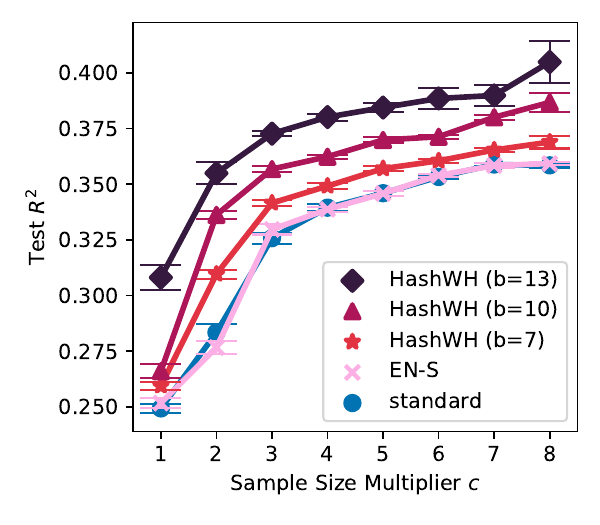}
         \caption{$n=100$, third draw of $g^*$}
     \end{subfigure}
    \caption{Generalization performance $R^2$ on a hold-out test set, in learning a synthetic degree $5$ function $g^*\in\{0,1\}^n$ for $n\in\{25,50,100\}$, using datasets of size $c \cdot 25n$. We report the results of the first draws of $g^*$ for each input dimension in Figure~\ref{fig:synthetic_large} and the extended version for all three draws of $g^*$ of different dimensions here. Our method, \textsc{HashWH}, \emph{always} outperforms the standard neural network and \textsc{EN-S}. We are capable of significantly increasing the outperformance margin by increasing $b$. \textsc{EN-S}, however, does not show improvement over the standard network in most cases which indicates its diminishing effectiveness as the size of the input dimension grows, i.e., the number of features increases.}
    \label{fig:full_synthetic_large}
\end{figure*}

Figure~\ref{fig:full_synthetic_large} shows the generalization performance of different methods in learning a synthetic degree $d=5$ function $g^*\in\{0,1\}^n\rightarrow\mathbb{R}$, for $n\in\{25,50,100\}$, using train sets of different sizes ($c\cdot25n, c\in[8]$). For each $n$ we sample three different draws of $g^*$. This is the extended version of Figure~\ref{fig:synthetic_large}, where we only reported the results for the first draw of $g^*$ for each input dimension $n$. Our regularization method, \textsc{HashWH}, outperforms the standard network and \textsc{EN-S} in all possible combinations of input dimension and dataset sizes, regardless of the draw of $g^*$. We observe that increasing $b$ in \textsc{HashWH}, i.e. increasing the number of hashing buckets, almost always improves the generalization performance. \textsc{EN-S}, on the other hand, does not show significant superiority over the standard neural network rather than marginally outperforming it in a few cases when $n=25$. This does not match its performance in the previous section and conveys that it is not able to perform well when increasing the input dimension, i.e., having more features in the data.

\begin{figure*}[h]
    \centering
    \begin{subfigure}[b]{0.49\linewidth}
         \centering
         \includegraphics[width=\linewidth]{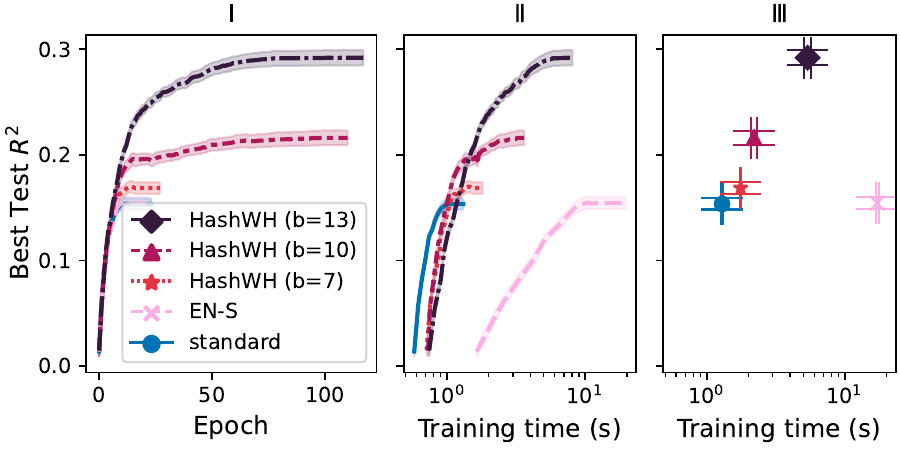}
         \caption{$n=25$, $c=1$}
     \end{subfigure}
    \begin{subfigure}[b]{0.49\linewidth}
         \centering
         \includegraphics[width=\linewidth]{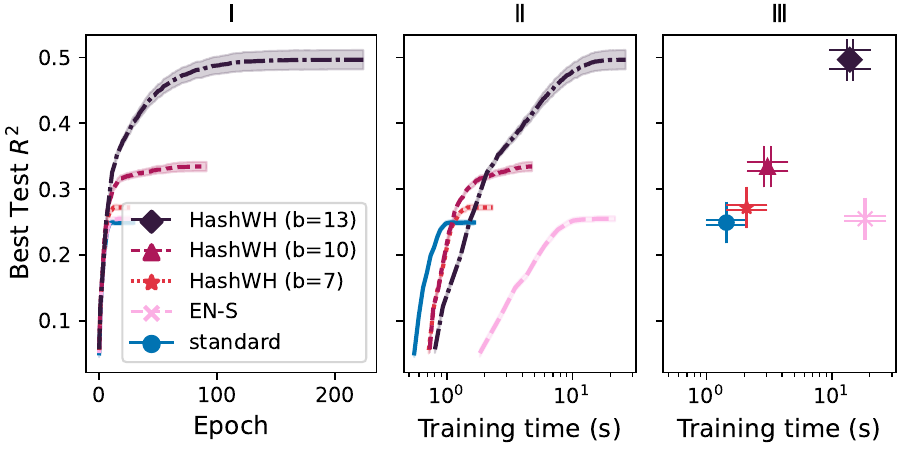}
         \caption{$n=25$, $c=2$}
     \end{subfigure}
    \begin{subfigure}[b]{0.49\linewidth}
         \centering
         \includegraphics[width=\linewidth]{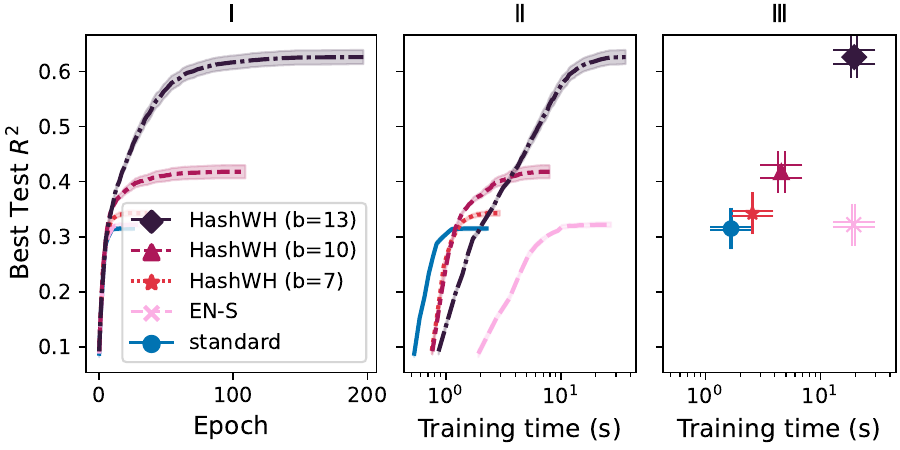}
         \caption{$n=25$, $c=3$}
     \end{subfigure}
    \begin{subfigure}[b]{0.49\linewidth}
         \centering
         \includegraphics[width=\linewidth]{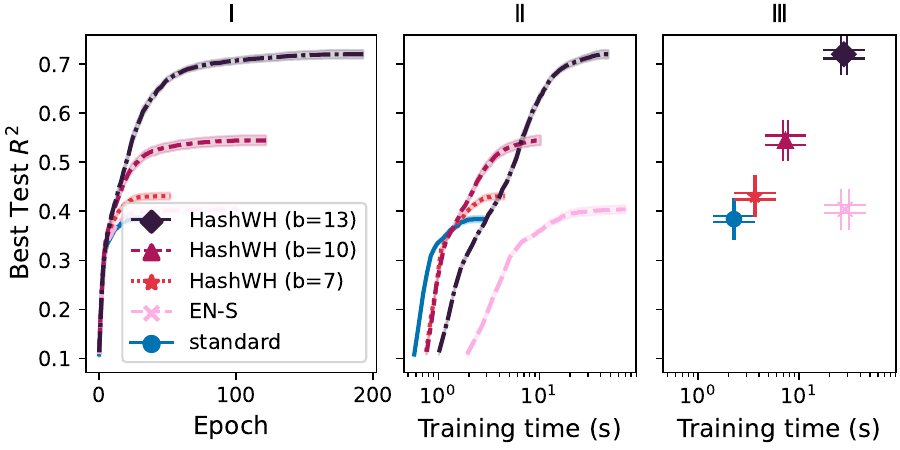}
         \caption{$n=25$, $c=4$}
     \end{subfigure}
    \begin{subfigure}[b]{0.49\linewidth}
         \centering
         \includegraphics[width=\linewidth]{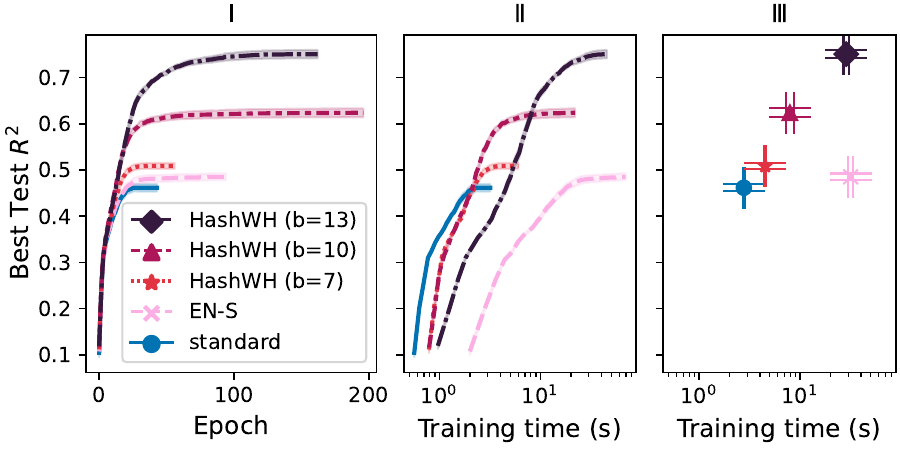}
         \caption{$n=25$, $c=5$}
     \end{subfigure}
    \begin{subfigure}[b]{0.49\linewidth}
         \centering
         \includegraphics[width=\linewidth]{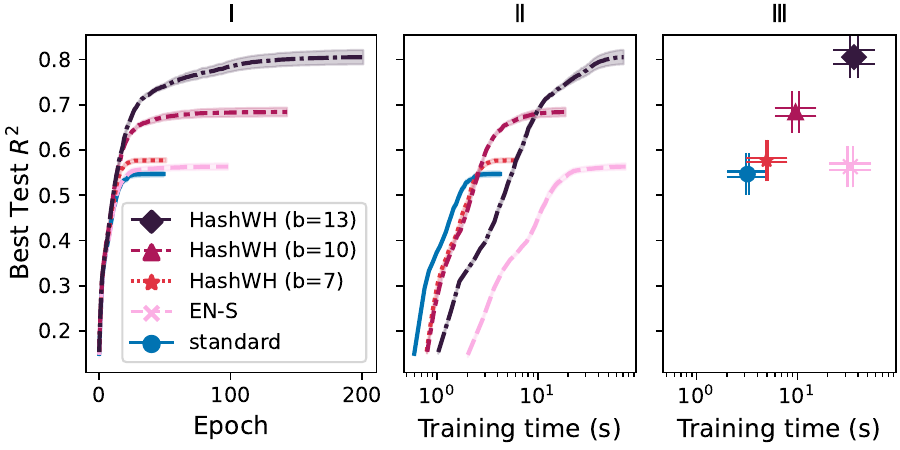}
         \caption{$n=25$, $c=6$}
     \end{subfigure}
    \begin{subfigure}[b]{0.49\linewidth}
         \centering
         \includegraphics[width=\linewidth]{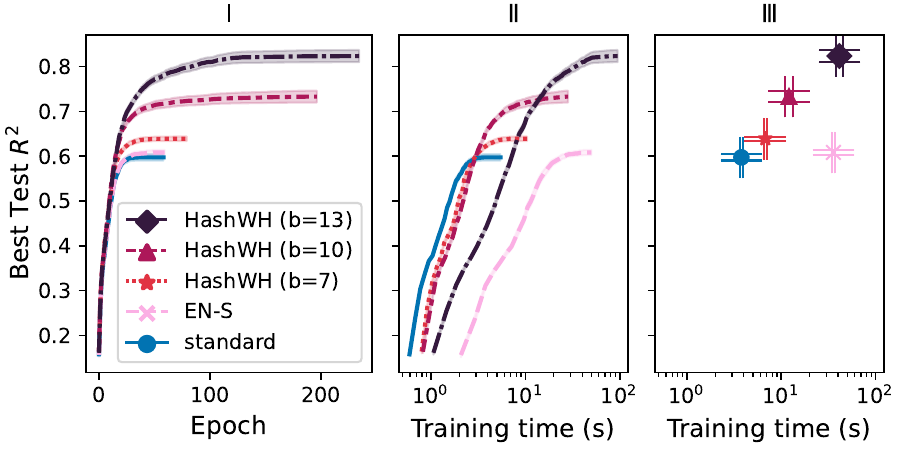}
         \caption{$n=25$, $c=7$}
     \end{subfigure}
    \begin{subfigure}[b]{0.49\linewidth}
         \centering
         \includegraphics[width=\linewidth]{plots/synthetic_large/triple_runtime_n25_size3_seed3.pdf}
         \caption{$n=25$, $c=8$}
     \end{subfigure}
    \caption{Best achievable generalization performance $R^2$ up to a certain epoch or training time (seconds), in learning a synthetic degree $5$ function $g^*\in\{0,1\}^n$, using datasets of size $c \cdot 25n$. This figure is an extended version of Figure~\ref{fig:runtime_synthetic_large}, where we reported similar plots for $n=50$ and $c=5$. Here we report the results for the first draw of $g^*$ with $n=25$. Our method, \textsc{HashWH}, \emph{always} outperforms \textsc{EN-S} $R^2$ score in significantly less time. \textsc{HashWH} can also be scaled by the choice of $b$ to achieve better generalization performance at th price of higher training times.}
    \label{fig:full25_runtime_synthetic_large}
\end{figure*}

\begin{figure*}[h]
    \centering
    \begin{subfigure}[b]{0.49\linewidth}
         \centering
         \includegraphics[width=\linewidth]{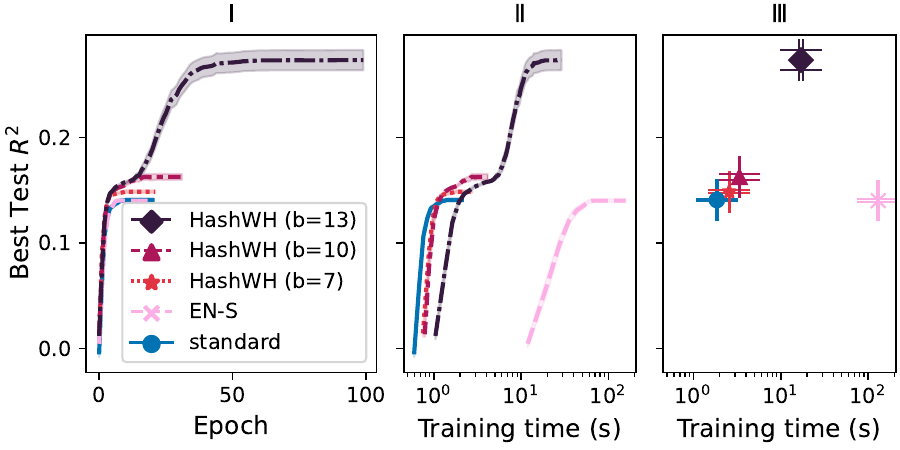}
         \caption{$n=50$, $c=1$}
     \end{subfigure}
    \begin{subfigure}[b]{0.49\linewidth}
         \centering
         \includegraphics[width=\linewidth]{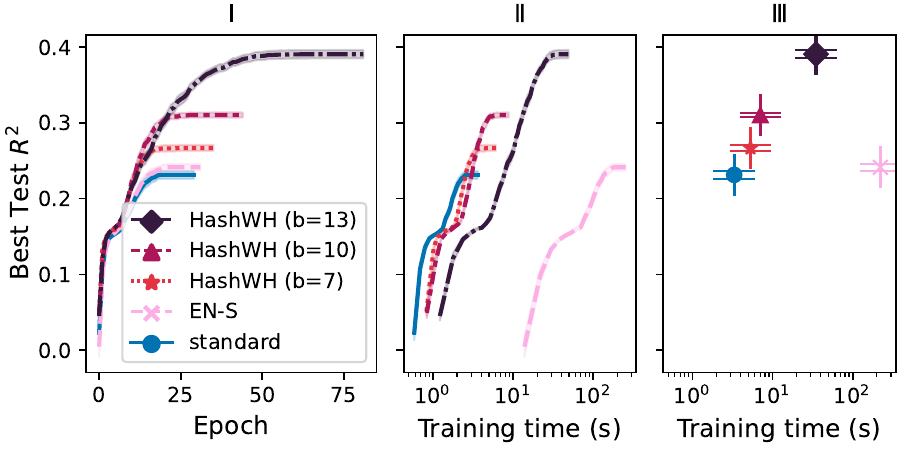}
         \caption{$n=50$, $c=2$}
     \end{subfigure}
    \begin{subfigure}[b]{0.49\linewidth}
         \centering
         \includegraphics[width=\linewidth]{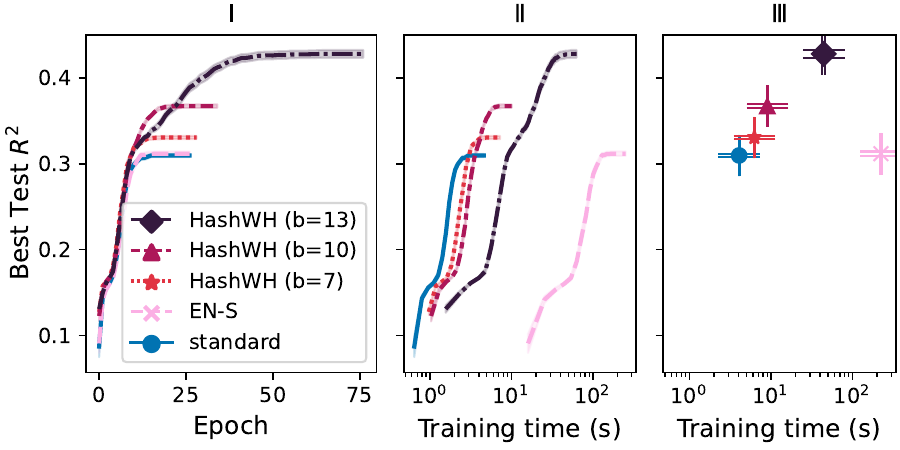}
         \caption{$n=50$, $c=3$}
     \end{subfigure}
    \begin{subfigure}[b]{0.49\linewidth}
         \centering
         \includegraphics[width=\linewidth]{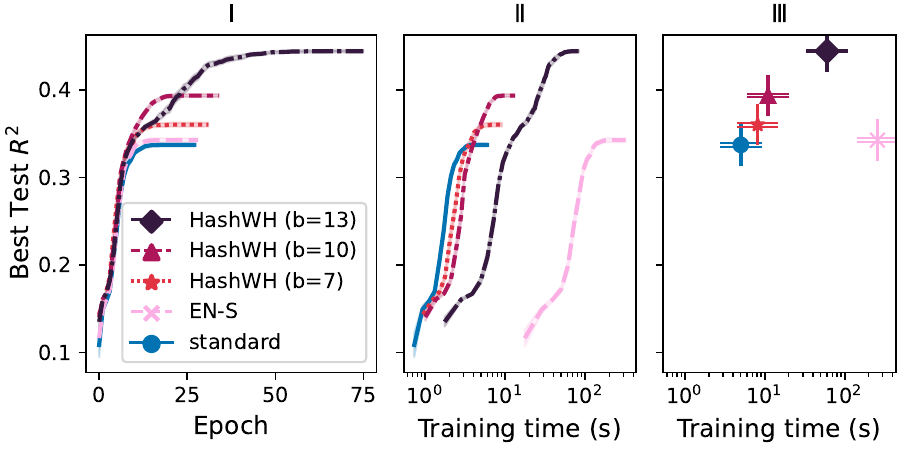}
         \caption{$n=50$, $c=4$}
     \end{subfigure}
    \begin{subfigure}[b]{0.49\linewidth}
         \centering
         \includegraphics[width=\linewidth]{plots/synthetic_large/triple_runtime_n50_size5_seed3.pdf}
         \caption{$n=50$, $c=5$}
     \end{subfigure}
    \begin{subfigure}[b]{0.49\linewidth}
         \centering
         \includegraphics[width=\linewidth]{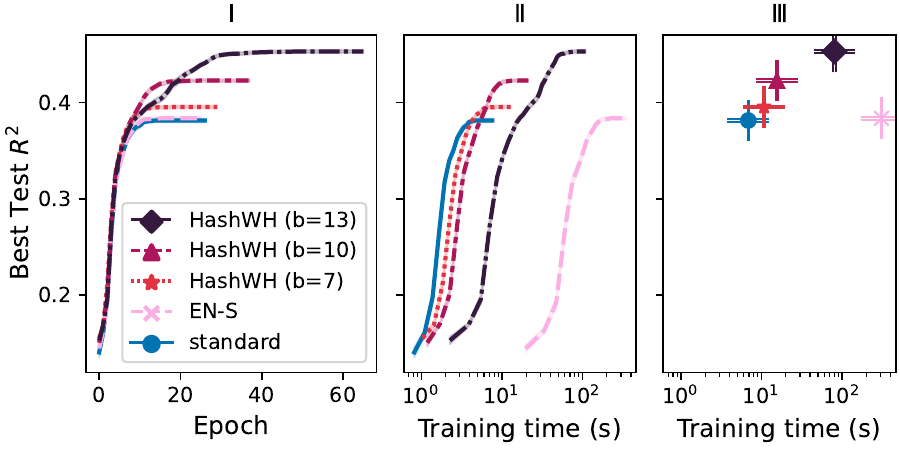}
         \caption{$n=50$, $c=6$}
     \end{subfigure}
    \begin{subfigure}[b]{0.49\linewidth}
         \centering
         \includegraphics[width=\linewidth]{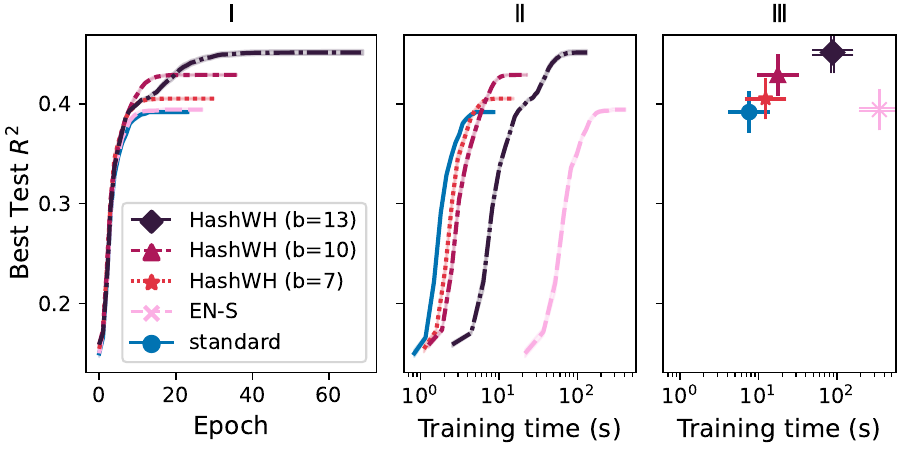}
         \caption{$n=50$, $c=7$}
     \end{subfigure}
    \begin{subfigure}[b]{0.49\linewidth}
         \centering
         \includegraphics[width=\linewidth]{plots/synthetic_large/triple_runtime_n50_size3_seed3.pdf}
         \caption{$n=50$, $c=8$}
     \end{subfigure}
    \caption{Best achievable generalization performance $R^2$ up to a certain epoch or training time (seconds), in learning a synthetic degree $5$ function $g^*\in\{0,1\}^n$, using datasets of size $c \cdot 25n$. This figure is an extended version of Figure~\ref{fig:runtime_synthetic_large}, where we reported similar plots for $n=50$ and $c=5$. Here we report the results for the first draw of $g^*$ with $n=50$. Our method, \textsc{HashWH}, \emph{always} outperforms \textsc{EN-S} $R^2$ score in significantly less time. \textsc{HashWH} can also be scaled by the choice of $b$ to achieve better generalization performance athe price of higher training times.}
    \label{fig:full50_runtime_synthetic_large}
\end{figure*}

\begin{figure*}[h]
    \centering
    \begin{subfigure}[b]{0.49\linewidth}
         \centering
         \includegraphics[width=\linewidth]{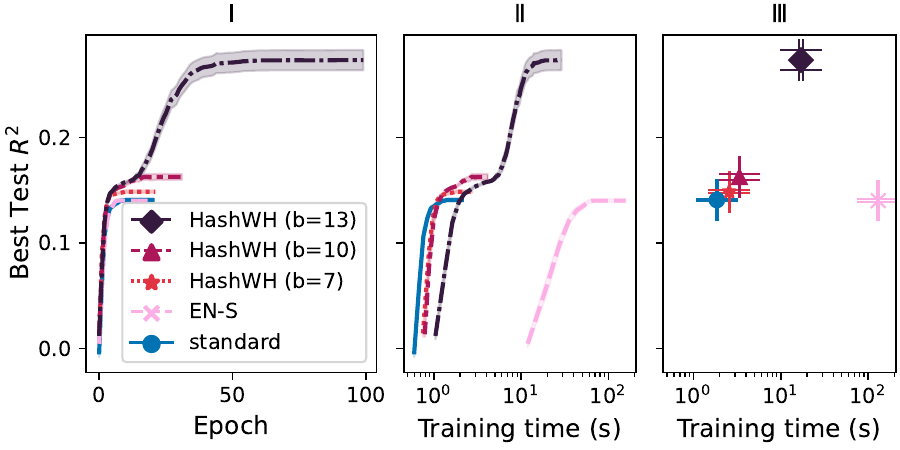}
         \caption{$n=100$, $c=1$}
     \end{subfigure}
    \begin{subfigure}[b]{0.49\linewidth}
         \centering
         \includegraphics[width=\linewidth]{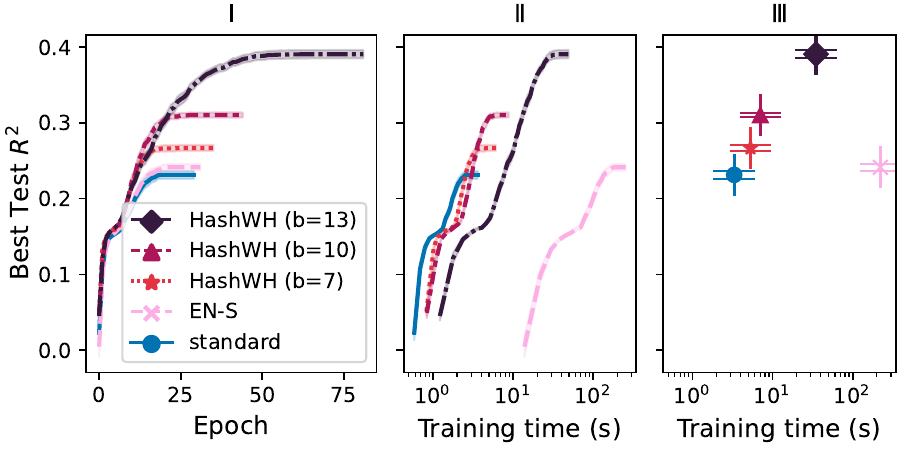}
         \caption{$n=100$, $c=2$}
     \end{subfigure}
    \begin{subfigure}[b]{0.49\linewidth}
         \centering
         \includegraphics[width=\linewidth]{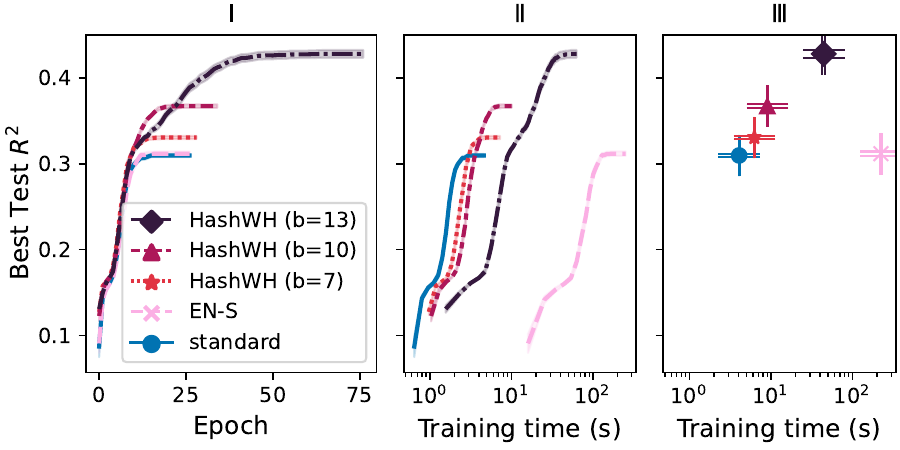}
         \caption{$n=100$, $c=3$}
     \end{subfigure}
    \begin{subfigure}[b]{0.49\linewidth}
         \centering
         \includegraphics[width=\linewidth]{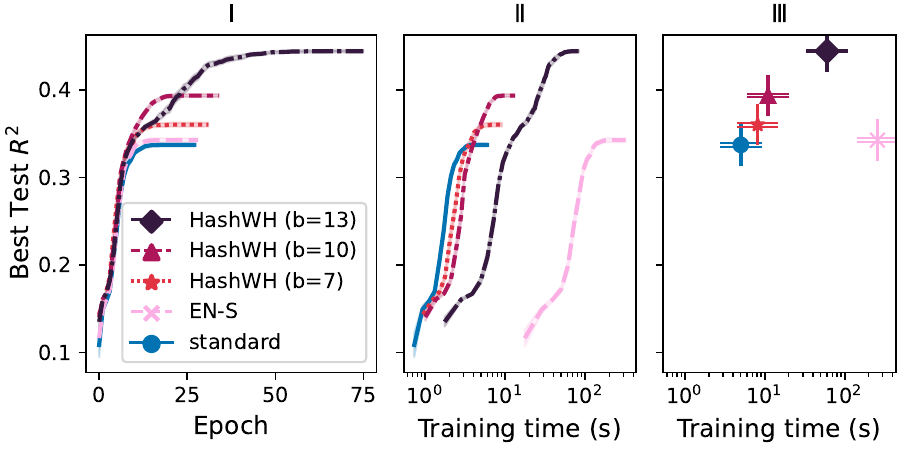}
         \caption{$n=100$, $c=4$}
     \end{subfigure}
    \begin{subfigure}[b]{0.49\linewidth}
         \centering
         \includegraphics[width=\linewidth]{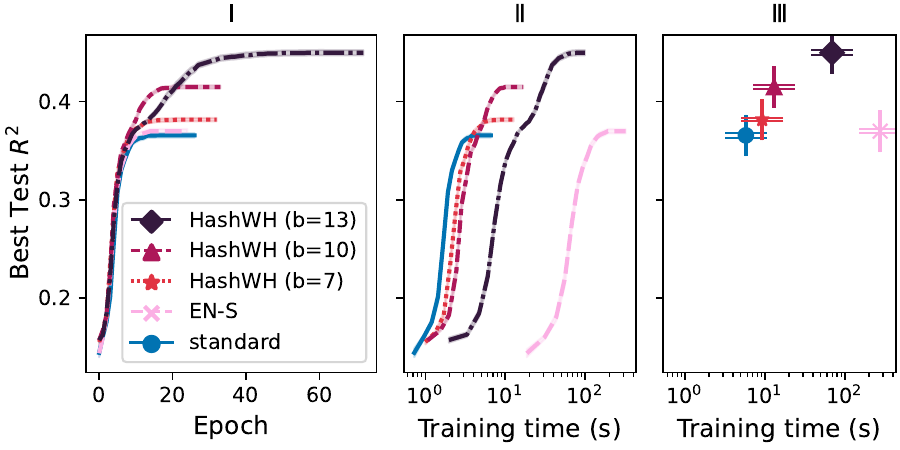}
         \caption{$n=100$, $c=5$}
     \end{subfigure}
    \begin{subfigure}[b]{0.49\linewidth}
         \centering
         \includegraphics[width=\linewidth]{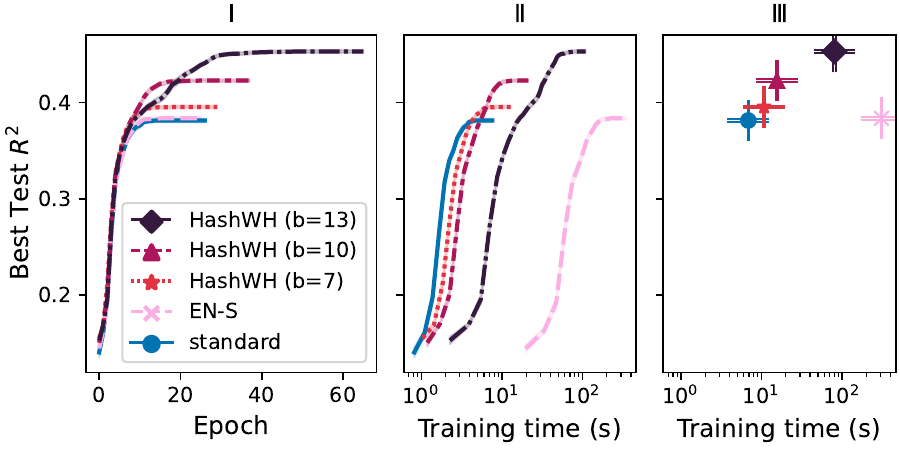}
         \caption{$n=100$, $c=6$}
     \end{subfigure}
    \begin{subfigure}[b]{0.49\linewidth}
         \centering
         \includegraphics[width=\linewidth]{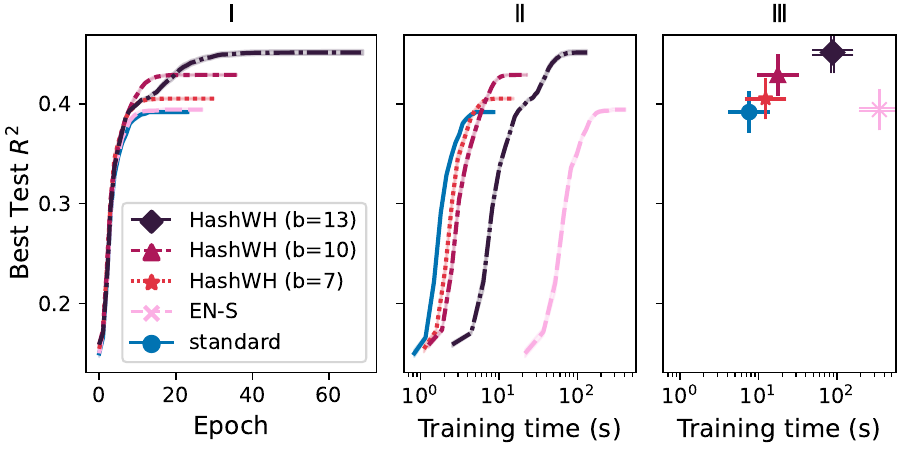}
         \caption{$n=100$, $c=7$}
     \end{subfigure}
    \begin{subfigure}[b]{0.49\linewidth}
         \centering
         \includegraphics[width=\linewidth]{plots/synthetic_large/triple_runtime_n100_size3_seed3.pdf}
         \caption{$n=100$, $c=8$}
     \end{subfigure}
    \caption{Best achievable generalization performance $R^2$ up to a certain epoch or training time (seconds), in learning a synthetic degree $5$ function $g^*\in\{0,1\}^n$, using datasets of size $c \cdot 25n$. This figure is an extended version of Figure~\ref{fig:runtime_synthetic_large}, where we reported similar plots for $n=50$ and $c=5$. Here we report the results for the first draw of $g^*$ with $n=100$. Our method, \textsc{HashWH}, \emph{always} outperforms \textsc{EN-S} $R^2$ score in significantly less time. \textsc{HashWH} can also be scaled by the choice of $b$ to achieve better generalization performance at the price of higher training times.}
    \label{fig:full100_runtime_synthetic_large}
\end{figure*}

To both showcase the computational scalability of our method, \textsc{HashWH}, and compare it to \textsc{EN-S}, we show the achievable performance by the number
of training epochs and training time in Figures~\labelcref{fig:full25_runtime_synthetic_large,fig:full50_runtime_synthetic_large,fig:full100_runtime_synthetic_large}, for all train set sizes and input dimensions individually and limited to the first draw of $g^*$ for each input dimension. This is the extended version of Figure~\ref{fig:runtime_synthetic_large} where we only reported it for $n=50$ and the sample size multiplier $c=5$. We consistently see that the trade-off between the generalization performance and the training time can be directly controlled in \textsc{HashWH} using the parameter $b$. Furthermore, \textsc{HashWH} is able to \emph{always} exhibit a significantly better generalization performance in remarkably less time, in all versions of $b$ tested. This emphasizes the advantage of our method in not directly computing the approximate Fourier spectrum of the network, which resulted in this gap with \textsc{EN-S} in the run time, that increases as the input dimension $n$ grows.

\subsection{Real data}
\label{app:subsec:real_detailed}
\begin{figure*}[h]
    \centering
     \begin{subfigure}[b]{0.65\linewidth}
         \centering
         \includegraphics[width=\linewidth]{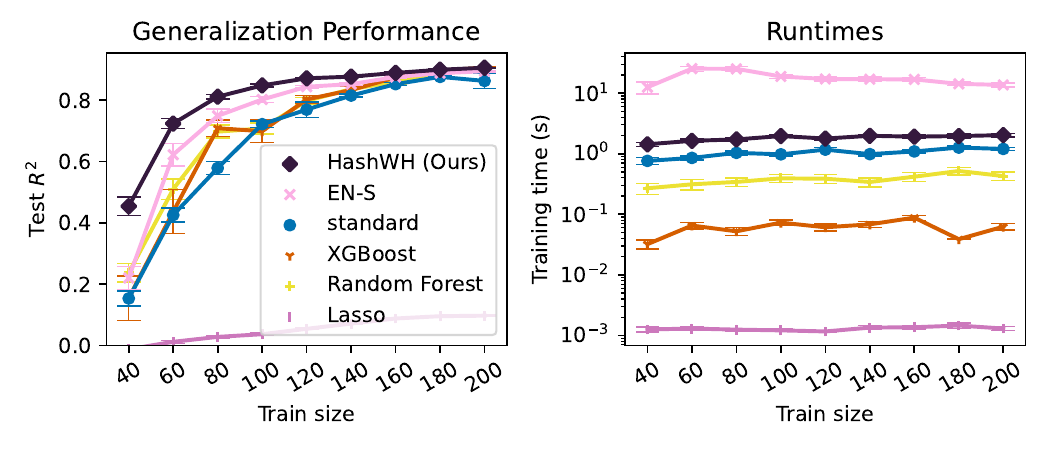}
         \caption{Entacmaea (n=13)}
     \end{subfigure}
     \begin{subfigure}[b]{0.65\linewidth}
         \centering
         \includegraphics[width=\linewidth]{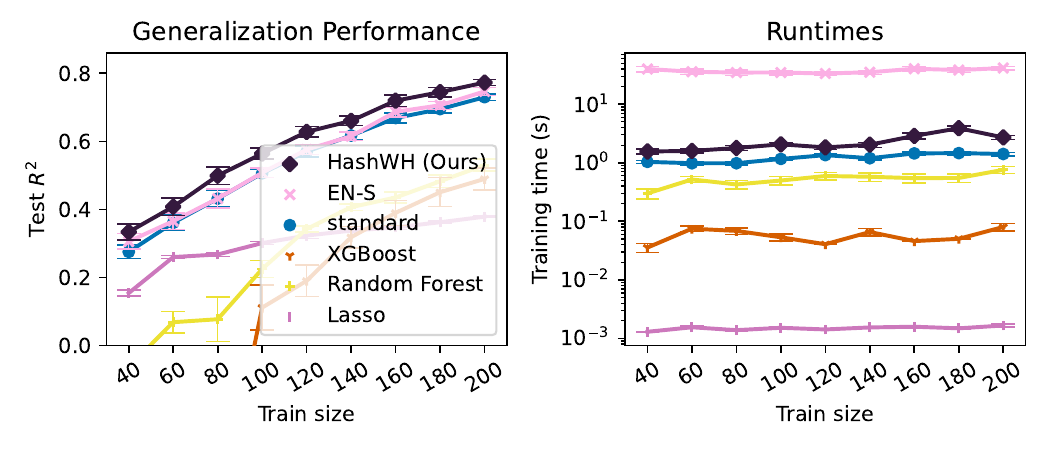}
         \caption{SGEMM (n=40)}
     \end{subfigure}
     \begin{subfigure}[b]{0.65\linewidth}
         \centering
         \includegraphics[width=\linewidth]{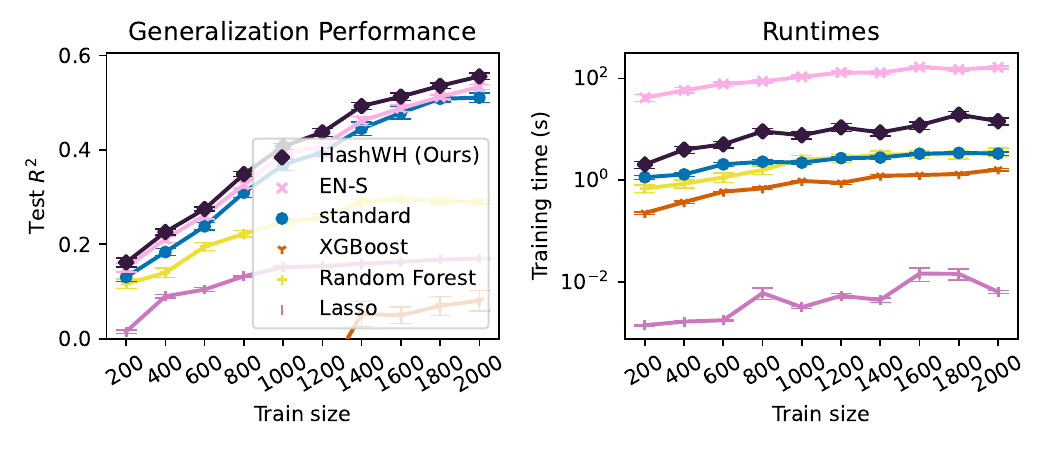}
         \caption{GB1 (n=80)}
     \end{subfigure}
     \begin{subfigure}[b]{0.65\linewidth}
         \centering
         \includegraphics[width=\linewidth]{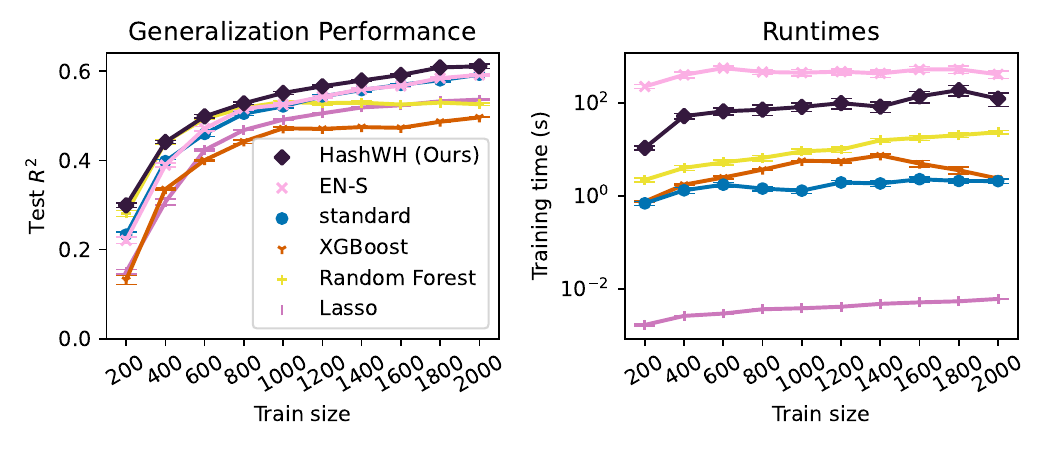}
         \caption{avGFP (n=236)}
    \end{subfigure}
    \caption{Generalization performance of standard/regularized neural networks and benchmark ML models on four real datasets. This figure is an extended version of Figure~\ref{fig:score_real_data}. It also includes the training times (logarithmically scaled in the plot). Our method is able to achieve the best test $R^2$s while always training significantly faster than \textsc{EN-S}.}
    \label{fig:full_performance_real_data}
\end{figure*}

Figure~\ref{fig:full_performance_real_data} shows the generalization performance and the training time of different methods, including relevant machine learning benchmarks, in learning four real datasets. It is the extended version of Figure~\ref{fig:score_real_data}, where we only reported the generalization performance and not the training time. The training time for neural nets is considered to be the time until overfitting occurs i.e. we do early stopping. \emph{In addition} to superior generalization performance of our method, \textsc{HashWH}, in most settings, again, we see that it is able to achieve it in significantly less time than \textsc{EN-S}. \textsc{Lasso} is the fastest among the methods but usually shows poor generalization performance.

\begin{figure*}[h]
  \centering
        \begin{subfigure}[b]{0.9\linewidth}
         \centering
            \includegraphics[width=\linewidth]{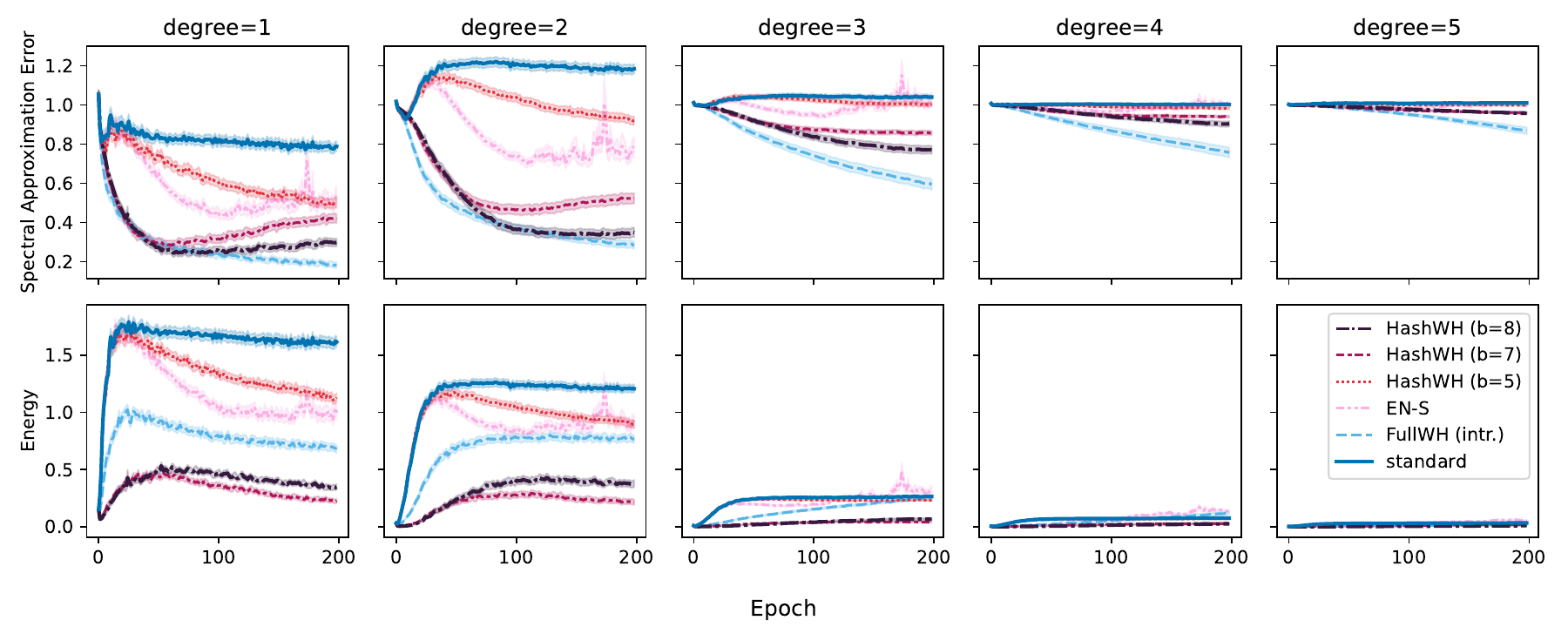}
         \caption{$\text{Train size}=100$}
     \end{subfigure}
     % \begin{subfigure}[b]{0.9\linewidth}
     %     \centering
     %        \includegraphics[width=\linewidth]{plots/spectrum/degree_split_function_error_n10_d5_size2.pdf}
     %     \caption{$\text{Train size}=200$}
     % \end{subfigure}
     \begin{subfigure}[b]{0.9\linewidth}
         \centering
            \includegraphics[width=\linewidth]{plots/spectrum/degree_split_function_error_n10_d5_size2.pdf}
         \caption{$\text{Train size}=200$}
     \end{subfigure}
     \begin{subfigure}[b]{0.9\linewidth}
         \centering
            \includegraphics[width=\linewidth]{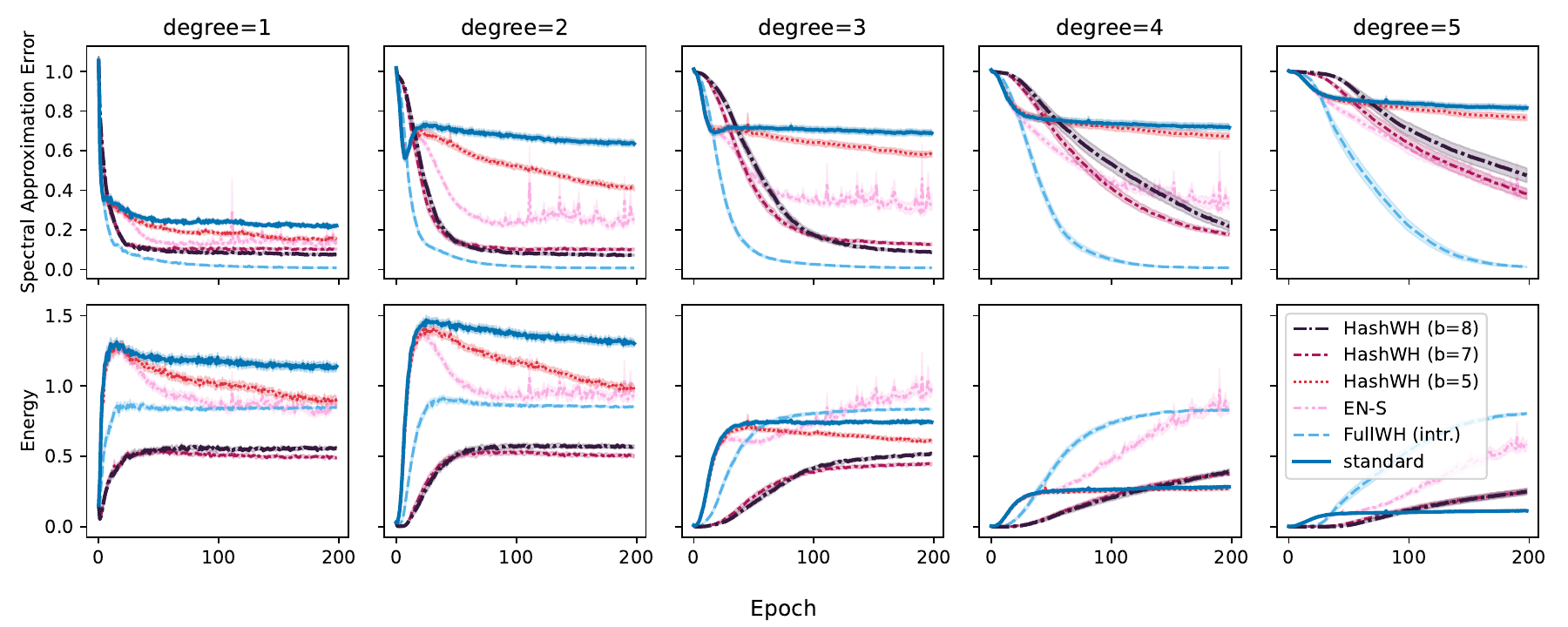}
         \caption{$\text{Train size}=300$}
     \end{subfigure}
  \caption{Evolution of the Spectral Approximation Error (SAE) and energy of the network during training, categorized by frequency degree (continued in the next page).}
\end{figure*}
\begin{figure*}[t!]\ContinuedFloat
    \centering
     \begin{subfigure}[b]{0.9\linewidth}
         \centering
            \includegraphics[width=\linewidth]{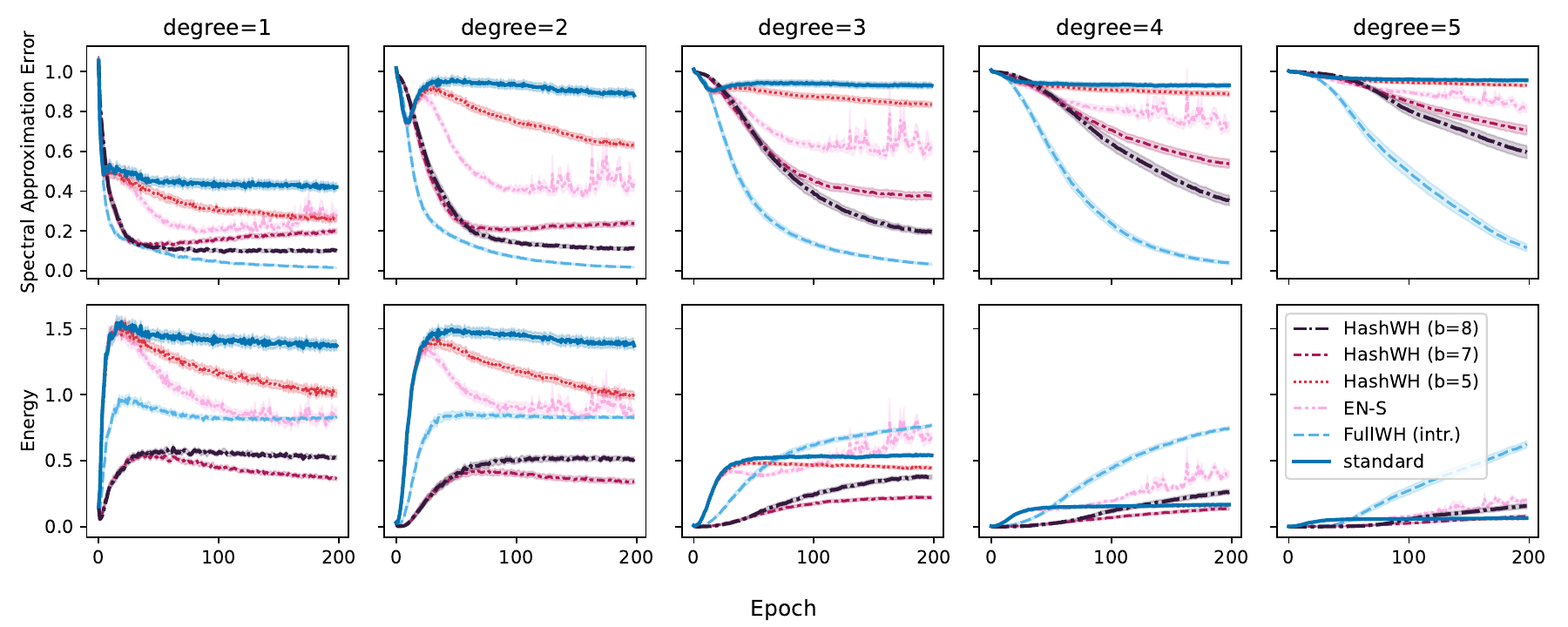}
         \caption{$\text{Train size}=400$}
     \end{subfigure}
      \caption{Evolution of the Spectral Approximation Error (SAE) and energy of the network during training, categorized by frequency degree. This is an extended version of Figure~\ref{fig:degree_split_function_error}, where we only reported results for training dataset size $200$. Firstly, in a standard neural network, the energy is mostly put on low-degree frequencies as compared to the high-degree frequencies. The energy slightly shifts towards high-degree frequencies when increasing the training dataset size. Our regularizer facilitates the learning of higher degrees in all cases. 
      Secondly, over the lower-degree and regardless of the train size, the standard neural network's energy continues to increase while the SAE first decreases then reverts and increases. This shows that the standard neural network emphasizes energy on erroneous low-degree frequencies and overfits. Our regularizer prevents overfitting in lower degrees.}
      \label{fig:full_degree_split_function_error}
\end{figure*}

% ---------------- High dimensional synthetic ----------------------

% --------------------- Real Data ----------------------
\end{document}